# The Communicative Multiagent Team Decision Problem: Analyzing Teamwork Theories and Models

**David V. Pynadath**                                                    PYNADATH@ISI.EDU
**Milind Tambe**                                                        TAMBE@USC.EDU
*Information Sciences Institute and Computer Science Department*
*University of Southern California*
*4676 Admiralty Way, Marina del Rey, CA 90292 USA*

## Abstract

Despite the significant progress in multiagent teamwork, existing research does not address the *optimality* of its prescriptions nor the *complexity* of the teamwork problem. Without a characterization of the optimality-complexity tradeoffs, it is impossible to determine whether the assumptions and approximations made by a particular theory gain enough efficiency to justify the losses in overall performance. To provide a tool for use by multiagent researchers in evaluating this tradeoff, we present a unified framework, the *COMmunicative Multiagent Team Decision Problem (COM-MTDP)*. The COM-MTDP model combines and extends existing multiagent theories, such as decentralized partially observable Markov decision processes and economic team theory. In addition to their generality of representation, COM-MTDPs also support the analysis of both the optimality of team performance and the computational complexity of the agents' decision problem. In analyzing complexity, we present a breakdown of the computational complexity of constructing optimal teams under various classes of problem domains, along the dimensions of observability and communication cost. In analyzing optimality, we exploit the COM-MTDP's ability to encode existing teamwork theories and models to encode two instantiations of joint intentions theory taken from the literature. Furthermore, the COM-MTDP model provides a basis for the development of novel team coordination algorithms. We derive a domain-independent criterion for optimal communication and provide a comparative analysis of the two joint intentions instantiations with respect to this optimal policy. We have implemented a reusable, domain-independent software package based on COM-MTDPs to analyze teamwork coordination strategies, and we demonstrate its use by encoding and evaluating the two joint intentions strategies within an example domain.

## 1. Introduction

A central challenge in the control and coordination of distributed agents is enabling them to work together, as a team, toward a common goal. Such teamwork is critical in a vast range of domains—for future teams of orbiting spacecraft, sensors for tracking targets, unmanned vehicles for urban battlefields, software agents for assisting organizations in rapid crisis response, etc. Research in teamwork theory has built the foundations for successful practical agent team implementations in such domains. On the forefront are theories based on belief-desire-intentions (BDI) frameworks, such as joint intentions (Cohen & Levesque, 1991b, 1991a; Levesque, Cohen, & Nunes, 1990), SharedPlans (Grosz, 1996; Grosz & Kraus, 1996; Grosz & Sidner, 1990), and others (Sonenberg, Tidhar, Werner, Kinny, Ljungberg, & Rao, 1994; Dunin-Keplicz & Verbrugge, 1996), that have provided prescriptions for co-ordination in practical systems. These theories have inspired the construction of practical, domain-independent teamwork models and architectures (Jennings, 1995; Pynadath, Tambe, Chauvat, & Cavedon, 1999; Rich & Sidner, 1997; Tambe, 1997; Yen, Yin, Ioerger, Miller, Xu, & Volz, 2001), successfully applied in a range of complex domains.

Yet, two key shortcomings limit the scalability of these BDI-based theories and implementations. First, there are no techniques for the quantitative evaluation of the degree of *optimality* of their coordination behavior. While optimal teamwork may be impractical in real-world domains, such analysis would aid us in comparison of different theories/models and in identifying feasible improvements. One key reason for the difficulty in quantitative evaluation of most existing teamwork theories is that they ignore the various uncertainties and costs in real-world environments. For instance, joint intentions theory (Cohen & Levesque, 1991b) prescribes that team members attain mutual beliefs in key circumstances, but it ignores the cost of attaining mutual belief (e.g., via communication). Implementations that blindly follow such prescriptions could engage in highly suboptimal coordination. On the other hand, practical systems have addressed costs and uncertainties of real-world environments. For instance, STEAM (Tambe, 1997; Tambe & Zhang, 1998) extends joint intentions with decision-theoretic communication selectivity. Unfortunately, the very pragmatism of such approaches often necessarily leads to a lack of theoretical rigor, so it remains unanswered whether STEAM's selectivity is the best an agent can do, or whether it is even necessary at all. The second key shortcoming of existing teamwork research is the lack of a characterization of the computational complexity of various aspects of teamwork decisions. Understanding the computational advantages of a practical coordination prescription could potentially justify the use of that prescription as an approximation to optimality in particular domains.

To address these shortcomings, we propose a new complementary framework, the *COMmunicative Multiagent Team Decision Problem (COM-MTDP)*, inspired by work in *economic team theory* (Marschak & Radner, 1971; Yoshikawa, 1978; Ho, 1980). While our COM-MTDP model borrows from a theory developed in another field, we make several contributions in applying and extending the original theory, most notably adding explicit models of communication and system dynamics. With these extensions, the COM-MTDP generalizes other recently developed multiagent decision frameworks, such as decentralized POMDPs (Bernstein, Zilberstein, & Immerman, 2000).

Our definition of a team (like that in economic team theory) assumes only that team members have a common goal and that they work selflessly towards that goal (i.e., they have no other private goals of their own). In terms of our decision-theoretic framework, we assume that all of the team members share the same joint utility function—that is, each team member's individual preferences are exactly the preferences of the other members and, thus, of the team as a whole. Our definition may appear to be a "bare-bones" definition of a team, since it does not include common concepts and assumptions from the literature on what constitutes a team (e.g., the teammates form a joint commitment (Cohen & Levesque, 1991b), attain mutual belief upon termination of a joint goal, intend that teammates succeed in their tasks (Grosz & Kraus, 1996), etc.). From our COM-MTDP perspective, we view these concepts as more intermediate concepts, as the *means* by which agents improve their team's overall performance, rather than ends in themselves. Our hypothesis in this investigation is that our COM-MTDP-based analysis can provide concrete justifications for these concepts. For example, while mutual belief has no inherent value, our COM-MTDP model can quantify the improved performance that we would expect from a team that attains mutual belief about important aspects of its execution.

More generally, this paper demonstrates three new types of teamwork analyses made possible by the COM-MTDP model. First, we analyze the computational complexity of teamwork within subclasses of problem domains. For instance, some researchers have advocated teamwork without communication (Goldberg & Mataric, 1997). We use the COM-MTDP model to show that, in general, the problem of constructing optimal teams without communication is NEXP-complete, but allowing free communication reduces the problem to be PSPACE-complete. This paper presents a breakdown of the complexity of optimal teamwork over problem domains classified along the dimensions of observability and communication cost.

Second, the COM-MTDP model provides a powerful tool for comparing the *optimality* of different coordination prescriptions across classes of domains. Indeed, we illustrate that we can encode existing team coordination strategies within a COM-MTDP for evaluation. For our analysis, we selected two joint intentions-based approaches from the literature: one using the approach realized within GRATE* and the joint responsibility model (Jennings, 1995), and another based on STEAM (Tambe, 1997). Through this encoding, we derive the conditions under which these team coordination strategies generate optimal team behavior, and the complexity of the decision problems addressed by them. Furthermore, we also derive a novel team coordination algorithm that outperforms these existing strategies in optimality, though not in efficiency. The end result is a *well-grounded characterization of the complexity-optimality tradeoff* among various means of team coordination.

Third, we can use the COM-MTDP model to empirically analyze a specific domain of interest. We have implemented reusable, domain-independent algorithms that allow one to evaluate the optimality of the behavior generated by different prescriptive policies within a problem domain represented as a COM-MTDP. We apply these algorithms in an example domain to empirically evaluate the aforementioned team coordination strategies, characterizing the optimality of each strategy as a function of the properties of the underlying domain. For instance, Jennings reports experimental results (Jennings, 1995) indicating that his joint responsibility teamwork model leads to lower waste of community effort than competing methods of accomplishing teamwork. With our COM-MTDP model, we were able to demonstrate the benefits of Jennings' approach under many configurations of our example domain. However, in precisely characterizing the types of domains that showed such benefits, we also identified domains where these competing methods may actually perform better. In addition, we can use our COM-MTDP model to re-create and explain previous work that noted an instance of suboptimality in a STEAM-based, real-world implementation (Tambe, 1997). While this previous work treated that suboptimality as anomalous, our COM-MTDP re-evaluation of the domain demonstrated that the observed suboptimality was a symptom of STEAM's general propensity towards extraneous communication in a significant range of domain types. Both the algorithms and the example domain model are available for public use in an Online Appendix 1.

Section 2 presents the COM-MTDP model's representation and places it in the context of related multiagent models from the literature. Section 3 uses the COM-MTDP model to define and characterize the complexity of designing optimal agent teams. Section 4 analyzes the optimality of existing team coordination algorithms and derives a novel coordination algorithm. Section 5 presents empirical results from applying our COM-MTDP algorithms to an example domain. Section 6 summarizes our results, and Section 7 identifies some promising future directions.

## 2. The COM-MTDP Model

This section defines and describes the COM-MTDP model itself and its ability to represent the important aspects of multiagent teamwork. We begin in Section 2.1 by defining the underlying multiagent team decision problem with no explicit communication. Section 2.2 defines the complete COM-MTDP model with its extension to explicitly represent communication. Section 2.3 provides an illustration of how the COM-MTDP model represents the execution of a team of agents. Finally, Section 2.4 describes related models of multiagent coordination and shows how the COM-MTDP model generalizes them.

### 2.1 Multiagent Team Decision Problems

Given a team of selfless agents, $\alpha$, who intend to perform some joint task, we wish to evaluate possible policies of behavior. We represent a *multiagent team decision problem* (MTDP) model as a tuple, $\langle S, \boldsymbol{A}_\alpha, P, \boldsymbol{\Omega}_\alpha, \boldsymbol{O}_\alpha, \boldsymbol{B}_\alpha, R \rangle$. We have taken the underlying components of this model from the initial team decision model (Ho, 1980), but we have extended them to handle dynamic decisions over time and to more easily represent multiagent domains (in particular, agent beliefs). We assume that the model is common knowledge to all of the team members. In other words, all of the agents believe the same model, and they believe that they all believe the same model, etc.

#### 2.1.1 WORLD STATES: $S$

- $S = \Xi_1 \times \cdots \times \Xi_m$: a set of world states, expressed as a factored representation (a cross product of separate features).

The state of the world here is the state of the team's environment (e.g., terrain, location of enemy). Thus, each $\Xi_i$ represents the domain of an individual feature of that environment, while $S$ represents the domain of all possible combinations of values over the individual features.

#### 2.1.2 DOMAIN-LEVEL ACTIONS: $\boldsymbol{A}_\alpha$

$\{A_i\}_{i \in \alpha}$ is a set of actions for each agent to perform to change its environment, implicitly defining a set of combined actions, $\boldsymbol{A}_\alpha \equiv \prod_{i \in \alpha} A_i$ (corresponding to team theory's *decision variables*).

**Extension to Dynamic Problem:** $P$   The original team decision problem focused on a one-shot, static problem. We extend the original concept so that each component is a time series of random variables. The effects of domain-level actions (e.g., a flying action changes a helicopter's position) obey a probabilistic distribution, given by a function $P : S \times \boldsymbol{A}_\alpha \times S \to [0, 1]$. In other words, for each initial state $s$ at time $t$, combined action $\boldsymbol{a}$ taken at time $t$, and final state $s'$ at time $t + 1$, $\Pr(S^{t+1} = s'|S^t = s, \boldsymbol{A}_\alpha^t = \boldsymbol{a}) = P(s, \boldsymbol{a}, s')$. The given definition of $P$ assumes that the world dynamics obey the Markov assumption.

### 2.1.3 Agent Observations: $\boldsymbol{\Omega}_\alpha$

$\{\Omega_i\}_{i \in \alpha}$ is a set of observations that each agent, $i$, can experience of its world, implicitly defining a combined observation, $\boldsymbol{\Omega}_\alpha \equiv \prod_{i \in \alpha} \Omega_i$. $\Omega_i$ may include elements corresponding to indirect evidence of the state (e.g., sensor readings) and actions of other agents (e.g., movement of other helicopters). In the original team-theoretic framework, the *information structure* that represented the observation process of the agents was a set of deterministic functions, $O_i : S \to \Omega_i$.

**Extension of Allowable Information Structures: $\boldsymbol{O}_\alpha$** We extend the information structure representation to allow for uncertain observations. We use a general stochastic model, borrowed from the *partially observable Markov decision process* model (Smallwood & Sondik, 1973), with a joint observation function: $\boldsymbol{O}_\alpha(s, \boldsymbol{a}, \boldsymbol{\omega}) = \Pr(\boldsymbol{\Omega}_\alpha^t = \boldsymbol{\omega}|S^t = s, \boldsymbol{A}_\alpha^{t-1} = \boldsymbol{a})$. This function models the sensors, representing any errors, noise, etc. In some cases, we can separate this joint distribution into individual observation functions: $\boldsymbol{O}_\alpha \equiv \prod_{i \in \alpha} O_i$, where $O_i(s, \boldsymbol{a}, \omega) = \Pr(\Omega_i^t = \omega|S^t = s, \boldsymbol{A}_\alpha^{t-1} = \boldsymbol{a})$. Thus, the probability distribution specified by $\boldsymbol{O}_\alpha$ forms the richer *information structure* used in our model. We can make useful distinctions between different classes of information structures:

**Collective Partial Observability** This is the general case, where we make no assumptions on the observations.

**Collective Observability** There is a unique world state for the *combined* observations of the team: $\forall \boldsymbol{\omega} \in \boldsymbol{\Omega}_\alpha$, $\exists s \in S$ such that $\forall s' \neq s$, $\Pr(\boldsymbol{\Omega}_\alpha^t = \boldsymbol{\omega}|S^t = s') = 0$. The set of domains that are collectively observable is a strict subset of the domains that are collectively partially observable.

**Individual Observability** There is a unique world state for each individual agent's observations: $\forall \omega \in \Omega_i$, $\exists s \in S$ such that $\forall s' \neq s$, $\Pr(\Omega_i^t = \omega|S^t = s') = 0$. The set of domains that are individually observable is a strict subset of the domains that are collectively observable.

**Non-Observability** The agents receive no feedback from the world: $\exists \omega \in \Omega_i$, such that $\forall s \in S$ and $\forall \boldsymbol{a} \in \boldsymbol{A}_\alpha$, $\Pr(\Omega_i^t = \omega|S^t = s, \boldsymbol{A}_\alpha^{t-1} = \boldsymbol{a}) = 1$. This assumption holds in open-loop systems, which come under frequent consideration in classical planning (Boutilier, Dean, & Hanks, 1999).

### 2.1.4 Policy (Strategy) Space

$\pi_{iA}$ is a domain-level *policy* (or *strategy*, in the original team theory specification) to map an agent's belief state to an action. In the original formalism, the agent's beliefs correspond directly to its observations (i.e., $\pi_{iA} : \Omega_i \to A_i$).

**Extension to Richer Belief State Space: $\boldsymbol{B}_\alpha$** We generalize the set of possible strategies to capture the more complex mental states of the agents. Each agent, $i \in \alpha$, forms a belief state, $b_i^t \in B_i$, based on its observations seen through time $t$, where $B_i$ circumscribes the set of possible belief states for the agent. Thus, we define the set of possible domain-level policies as mappings from belief states to actions, $\pi_{iA} : B_i \rightarrow A_i$. We define the set of possible combined belief states over all agents to be $\boldsymbol{B}_\alpha \equiv \prod_{i \in \alpha} B_i$. The corresponding random variable, $\boldsymbol{b}_\alpha^t$, represents the agents' combined belief state at time $t$. We elaborate on different types of belief states and the mapping of observations to belief states (i.e., the *state estimator function*) in Section 2.2.1.

### 2.1.5 REWARD FUNCTION: $R$

A common reward function is central to the notion of teamwork in a MTDP: $R : S \times \boldsymbol{A}_\alpha \rightarrow \mathbb{R}$. This function represents the team's joint preferences over states and the cost of domain-level actions (e.g., destroying enemy is good, returning to home base with only 10% of original force is bad). We assume that, as selfless team members, each agent shares these preferences at the individual level as well. Therefore, each team member wants exactly what is best for the team as a whole.

## 2.2 Extension for Explicit Communication: $\Sigma_\alpha$

We make an explicit separation between domain-level actions ($\boldsymbol{A}_\alpha$) and communicative actions. As defined in this section, communicative actions affect the receiving agents' individual belief states, but, unlike domain-level actions, they do not directly change the world state. Although this distinction is sometimes blurry in real-world domains, we make this explicit separation so as to isolate, as much as possible, the effects of the two types of actions. The leverage gained from this separation provides the basis for the informative, analytical results presented in the rest of this paper. To capture this separation, we extend our initial MTDP model to be a *communicative multiagent team decision problem* (COM-MTDP), that we define as a tuple, $\langle S, \boldsymbol{A}_\alpha, \boldsymbol{\Sigma}_\alpha, P, \boldsymbol{\Omega}_\alpha, \boldsymbol{O}_\alpha, \boldsymbol{B}_\alpha, R \rangle$, with a new component, $\boldsymbol{\Sigma}_\alpha$, and an extended reward function, $R$.

### 2.2.1 COMMUNICATION: $\boldsymbol{\Sigma}_\alpha$

$\{\Sigma_i\}_{i \in \alpha}$ is a set of possible messages for each agent, implicitly defining a set of combined communications, $\boldsymbol{\Sigma}_\alpha \equiv \prod_{i \in \alpha} \Sigma_i$. An agent, $i$, may communicate message $x \in \Sigma_i$ to its teammates, who interpret the communication by updating their belief states in response. As a first step in this work, we assume that all of the agents receive the messages instantaneously and correctly (i.e., there is no lag or noise in the communication channels). This model is common knowledge among all of the team members, so once an agent has sent a message, it knows that its team members have received the message, and its team members know that it knows that they have all received the message, and so on.

With communication, we divide each decision epoch into two phases: the *pre-communication* and *post-communication* phases, denoted by the subscripts $\bullet\Sigma$ and $\Sigma\bullet$, respectively. In particular, the agents update their belief states at two distinct points within each decision epoch: once upon receiving observation $\Omega_i^t$ (producing the pre-communication belief state $b_{i\bullet\Sigma}^t$), and again upon receiving the other agents' messages (producing the post-communication belief state $b_{i\Sigma\bullet}^t$). The distinction allows us to differentiate between the belief state used by the agents in selecting their communication actions and the more "up-to-date" belief state used in selecting their domain-level actions. We also distinguish between the separate *state-estimator* functions used in each update phase:

$$b_i^0 = SE_i^0() \tag{1}$$

$$b_{i\bullet\Sigma}^t = SE_{i\bullet\Sigma}(b_{i\Sigma\bullet}^{t-1}, \Omega_i^t) \tag{2}$$

$$b_{i\Sigma\bullet}^t = SE_{i\Sigma\bullet}(b_{i\bullet\Sigma}^t, \mathbf{\Sigma}_\alpha^t) \tag{3}$$

where $SE_{i\bullet\Sigma} : B_i \times \Omega_i \to B_i$ is the pre-communication state estimator for agent $i$, and $SE_{i\Sigma\bullet} : B_i \times \mathbf{\Sigma}_\alpha \to B_i$ is the post-communication state estimator for agent $i$. The initial state estimator, $SE_i^0 : \emptyset \to B_i$, specifies the agent's prior beliefs, before any observations are made. For each of these, we also make the obvious definitions for the corresponding estimators for the combined belief states: $\boldsymbol{SE}_{\alpha\bullet\Sigma}$, $\boldsymbol{SE}_{\alpha\Sigma\bullet}$, and $\boldsymbol{SE}_\alpha^0$.

In this paper, *as a first step*, we assume that the agents have *perfect recall*. In other words, the agents recall all of their observations, as well as all communication of the other agents. Thus, their belief states can represent their entire histories as sequences of observations and received messages: $B_i = \Omega_i^* \times \mathbf{\Sigma}_\alpha^*$, where $X^*$ denotes the set of all possible sequences (of any length) of elements of $X$. The agents realize perfect recall through the following state estimator functions:

$$SE_i^0() = \langle\rangle \tag{4}$$

$$SE_{i\bullet\Sigma}(\langle\langle\Omega_i^0, \mathbf{\Sigma}_\alpha^0\rangle, \dots, \langle\Omega_i^{t-1}, \mathbf{\Sigma}_\alpha^{t-1}\rangle\rangle, \Omega_i^t)$$
$$= \langle\langle\Omega_i^0, \mathbf{\Sigma}_\alpha^0\rangle, \dots, \langle\Omega_i^{t-1}, \mathbf{\Sigma}_\alpha^{t-1}\rangle, \langle\Omega_i^t, \cdot\rangle\rangle \tag{5}$$

$$SE_{i\Sigma\bullet}(\langle\langle\Omega_i^0, \mathbf{\Sigma}_\alpha^0\rangle, \dots, \langle\Omega_i^{t-1}, \mathbf{\Sigma}_\alpha^{t-1}\rangle, \langle\Omega_i^t, \cdot\rangle\rangle, \mathbf{\Sigma}_\alpha^t)$$
$$= \langle\langle\Omega_i^0, \mathbf{\Sigma}_\alpha^0\rangle, \dots, \langle\Omega_i^t, \mathbf{\Sigma}_\alpha^t\rangle\rangle \tag{6}$$

In other words, $SE_i^0$ initializes agent $i$'s belief state to be an empty history, $SE_{i\bullet\Sigma}$ appends a new observation to agent $i$'s belief state, and $SE_{i\Sigma\bullet}$ appends new messages to agent $i$'s belief state. Under this paper's assumptions of perfect recall, all three state-estimator functions take only constant time. However, we can potentially allow more complex functions (though the complexity results presented hold only if the state-estimator functions take polynomial time). For instance, although we assume perfect, synchronous, instantaneous communication here, we could potentially use the post-communication state estimator to model any noise, temporal delays, asynchrony, cognitive burden, etc. present in the communication channel.

We extend our definition of a policy of behavior to include a *communication policy*, $\pi_{i\Sigma} : B_i \to \Sigma_i$, analogous to Section 2.1.4's domain-level policy. We define the joint policies, $\boldsymbol{\pi}_{\alpha\Sigma}$ and $\boldsymbol{\pi}_{\alpha A}$, as the combined policies across all agents in $\alpha$.

### 2.2.2 Extended Reward Function: $R$

We extend the team's reward function to also represent the cost of communicative acts (e.g., communication channels may have associated cost): $R : S \times \boldsymbol{A}_\alpha \times \boldsymbol{\Sigma}_\alpha \to \mathbb{R}$. We assume that the cost of communication and of domain-level actions are independent of each other, so we can decompose the reward function into two components: a communication-level reward, $R_\Sigma : S \times \boldsymbol{\Sigma}_\alpha \to \mathbb{R}$, and a domain-level reward, $R_A : S \times \boldsymbol{A}_\alpha \to \mathbb{R}$. The total reward is the sum of the two component values: $R(s, \boldsymbol{a}, \boldsymbol{\sigma}) = R_A(s, \boldsymbol{a}) + R_\Sigma(s, \boldsymbol{\sigma})$. We assume that communication has no inherent benefit and may instead have some cost, so that for all states, $s \in S$, and messages, $\boldsymbol{\sigma} \in \boldsymbol{\Sigma}_\alpha$, the reward is never positive: $R_\Sigma(s, \boldsymbol{\sigma}) \leq 0$. However, although we assign communication no explicit value, it can have significant implicit value through its effect on the agents' belief states and, subsequently, on their future actions.

As with the observability function, we parameterize the communication costs associated with message transmissions:

**General Communication:** We make no assumptions about communication.

**Free Communication:** $R_\Sigma(s, \boldsymbol{\sigma}) = 0$ for any $\boldsymbol{\sigma} \in \boldsymbol{\Sigma}_\alpha$, and $s \in S$. In other words, communication actions have no effect on the agents' reward.

**No communication:** $\boldsymbol{\Sigma}_\alpha = \emptyset$, i.e., no *explicit* communication. Alternatively, communication may be prohibitively expensive, so that $\forall \boldsymbol{\sigma} \in \boldsymbol{\Sigma}_\alpha$, and $s \in S$, $R_\Sigma(s, \boldsymbol{\sigma}) = -\infty$.

The *free-communication* case appears in the literature, when researchers wish to focus on issues other than communication cost. Although, real-world domains rarely exhibit such ideal conditions, we may be able to model some domains as having approximately free communication to a sufficient degree. In addition, analyzing this extreme case gives us some understanding of the benefit of communication, even if the results do not apply across all domains. We also identify the *no-communication* case because such decision problems have been of interest to researchers as well (Goldberg & Mataric, 1997). Of course, even if $\boldsymbol{\Sigma}_\alpha = \emptyset$, it is possible that there are domain-level actions in $\boldsymbol{A}_\alpha$ that have *implicit* communicative value by acting as signals that convey information to the other agents. However, we still label such agent teams as having *no communication* for the purposes of the work here, since many of our results exploit an *explicit* separation between domain- and communication-level actions.

## 2.3 Model Illustration

We can view the evolving state as a Markov chain with separate stages for domain-level and communication-level actions. In other words, each agent team member, $i \in \alpha$ begins in some initial state, $S^0$, with initial belief states, $b_i^0 = SE_i^0()$. Each agent receives an observation $\Omega_i^0$ drawn according to the probability distribution $\boldsymbol{O}_\alpha(S^0, \text{null}, \boldsymbol{\Omega}_\alpha^0)$ (there are no actions yet). Then, each agent updates its belief state, $b_{i\bullet\Sigma}^0 = SE_{i\bullet\Sigma}(b_i^0, \Omega_i^0)$.

Next, each agent $i \in \alpha$ selects a message according to its communication policy, $\Sigma_i^0 = \pi_{i\Sigma}(b_{i\bullet\Sigma}^0)$, defining a combined communication, $\boldsymbol{\Sigma}_\alpha^0$. Each agent interprets the communications of all of the others by updating its belief state, $b_{i\Sigma\bullet}^0 = SE_{i\Sigma\bullet}(b_{i\bullet\Sigma}^0, \boldsymbol{\Sigma}_\alpha^0)$. Each then selects an action according to its domain-level policy, $A_i^0 = \pi_{iA}(b_{i\Sigma\bullet}^0)$, defining a combined action $\boldsymbol{A}_\alpha^0$. By our central assumption of teamwork, each agent receives the same joint reward, $R^0 = R(S^0, \boldsymbol{A}_\alpha^0, \boldsymbol{\Sigma}_\alpha^0)$. The world then moves into a new state, $S^1$, according to the distribution, $P(S^0, \boldsymbol{A}_\alpha^0)$. Again, each agent $i$ receives an observation $\Omega_i^1$ drawn from $\Omega_i$ according to the distribution $\boldsymbol{O}_\alpha(S^1, \boldsymbol{A}_\alpha^0, \boldsymbol{\Omega}_\alpha^1)$, and it updates its belief state, $b_{i\bullet\Sigma}^1 = SE_{i\bullet\Sigma}(b_{i\Sigma\bullet}^0, \Omega_i^1)$.

The process continues, with agents choosing communication- and domain-level actions, observing the effects, and updating their beliefs. Thus, in addition to the time series of world states, $S^0, S^1, \ldots, S^t$, the agents themselves determine a time series of communication-level and domain-level actions, $\boldsymbol{\Sigma}_\alpha^0, \boldsymbol{\Sigma}_\alpha^1, \ldots, \boldsymbol{\Sigma}_\alpha^t$ and $\mathbf{A}_\alpha^1, \mathbf{A}_\alpha^1, \ldots, \mathbf{A}_\alpha^t$, respectively. We also have a time series of observations for each agent $i$, $\Omega_i^0, \Omega_i^1, \ldots, \Omega_i^t$. Likewise, we can treat the combined observations, $\boldsymbol{\Omega}_\alpha^0, \boldsymbol{\Omega}_\alpha^1, \ldots, \boldsymbol{\Omega}_\alpha^t$, as a similar time series of random variables.

Finally, the agents receive a series of rewards, $R^0, R^1, \ldots, R^t$. We can define the *value*, $V$, of the policies, $\boldsymbol{\pi}_{\alpha A}$ and $\boldsymbol{\pi}_{\alpha \Sigma}$, as the expected reward received when executing those policies. Over a finite horizon, $T$, this value is equivalent to the following:

$$V^T(\boldsymbol{\pi}_{\alpha A}, \boldsymbol{\pi}_{\alpha \Sigma}) = E\left[\sum_{t=0}^{T} R^t \,\middle|\, \boldsymbol{\pi}_{\alpha A}, \boldsymbol{\pi}_{\alpha \Sigma}\right] \tag{7}$$

## 2.4 Related Work

The COM-MTDP model subsumes many existing multiagent models, as presented in Table 1 (i.e., we can map any instance of these models into a corresponding COM-MTDP). This generality enables us to perform novel analyses of real-world teamwork domains, as demonstrated by Section 4's use of the COM-MTDP model for analyzing the optimality of communication decisions.

### 2.4.1 DECENTRALIZED POMDPs

With its model of observability and world dynamics, our COM-MTDP model closely parallels the structure of the *decentralized partially observable Markov decision process* (DEC-POMDP) (Bernstein et al., 2000). Following our notational conventions, a DEC-POMDP is a tuple, $\langle S, \boldsymbol{A}_\alpha, P, \boldsymbol{\Omega}_\alpha, \ \boldsymbol{O}_\alpha, R \rangle$. There is no set of possible messages, $\boldsymbol{\Sigma}_\alpha$, so the DEC-POMDP falls into the class of domains with *no communication*. The DEC-POMDP observational model, $\mathbf{O}$, is general enough to capture *collectively partially observable* domains.

### 2.4.2 PARTIALLY OBSERVABLE IDENTICAL PAYOFF STOCHASTIC GAMES

Stochastic games provide a rich framework for multiagent decision making when the agents may have their own individual goals and preferences. The *identical payoff stochastic game* (IPSG) restricts the agents to share a single payoff function, appropriate for modeling the single, global reward function of the team context. The *partially observable IPSG* (POIPSG) (Peshkin, Kim, Meuleau, & Kaelbling, 2000) is a tuple, $\langle S, \boldsymbol{A}_\alpha, P, \boldsymbol{\Omega}_\alpha, \boldsymbol{O}_\alpha, R \rangle$, very similar to the DEC-POMDP model. In other words, the observation function, $\boldsymbol{O}_\alpha$, is general enough to support *collectively partially observable* domains, and there is *no communication*.

### 2.4.3 MULTIAGENT MDPs

Another relevant model is the *multiagent Markov decision process* (MMDP) (Boutilier, 1996), which is a tuple, $\langle S, \boldsymbol{A}_\alpha, P, R \rangle$, in our notation. Like the DEC-POMDP, the MMDP has *no communication*. In addition, the MMDP is a multiagent extension to the completely observable MDP model, so it assumes an environment that is *individually observable*.

| Model | $\Sigma_\alpha$ | $O_\alpha$ |
|---|---|---|
| DEC-POMDP | no communication | collective partial observability |
| POIPSG | no communication | collective partial observability |
| MMDP | no communication | individual observability |
| Xuan-Lesser | general communication | collective observability |

Table 1: Existing models as COM-MTDP subsets.

### 2.4.4 XUAN-LESSER FRAMEWORK

The COM-MTDP's separation of communication from other actions is similar to previous work on multiagent decision models (Xuan, Lesser, & Zilberstein, 2001), which supported *general communication*. However, while the Xuan-Lesser model generalizes beyond individually observable environments, it supports only a subset of *collectively observable* environments. In particular, the Xuan-Lesser framework cannot represent agents who receive local observations of a common world state, where the observations of different agents could potentially be interdependent.

## 3. COM-MTDP Complexity Analysis

We can use the COM-MTDP model to prove some results about the complexity of constructing optimal agent teams (i.e., teams that coordinate to produce optimal behavior in a problem domain). The problem facing these agents (or the designer of these agents) is how to construct the joint policies, $\boldsymbol{\pi}_{\alpha\Sigma}$ and $\boldsymbol{\pi}_{\alpha A}$, so as to maximize their joint reward, as represented by the expected value, $V^T(\boldsymbol{\pi}_{\alpha A}, \boldsymbol{\pi}_{\alpha\Sigma})$. In all of the results presented, we assume that all of the values in a model instance (e.g., transition probabilities, rewards) are rational numbers, so that we can express the particular instance as a finite-sized input.

**Theorem 1** *The decision problem of whether there exist policies, $\boldsymbol{\pi}_{\alpha\Sigma}$ and $\boldsymbol{\pi}_{\alpha A}$, for a given COM-MTDP, under* general communication *and* collective partial observability, *that yield a total reward at least $K$ over some finite horizon $T$ is NEXP-complete if $|\alpha| \geq 2$ (i.e., more than one agent).*

**Proof:** To prove that the COM-MTDP decision problem is NEXP-hard, we reduce a DEC-POMDP (Bernstein et al., 2000) to a COM-MTDP with no communication by copying all of the other model features from the given DEC-POMDP. In other words, if we are given a DEC-POMDP, $\langle S, \{A^i\}_{i=1}^m, P, \{\Omega^i\}_{i=1}^m, O, R \rangle$, we can construct a COM-MTDP, $\langle S', \{A_i'\}_{i=1}^m, \boldsymbol{\Sigma}_\alpha', P', \{\Omega_i'\}_{i=1}^m, \boldsymbol{O}_\alpha', \boldsymbol{B}_\alpha', R' \rangle$, as follows:

$S' = S$

$A_i' = A^i$

$\Sigma' = \emptyset$

$P'(s, \langle a_1, \ldots, a_m \rangle, s') = P(s'|s, a_1, \ldots, a_m)$

$$\Omega_i' = \Omega^i$$

$$\boldsymbol{O}_\alpha'(s, \langle a_1, \dots, a_m \rangle, \langle \omega_1, \dots, \omega_m \rangle) = O(\omega_1, \dots, \omega_m | a_1, \dots, a_m, s)$$

$$B_i' = \cup_{j=1}^T (\Omega^i)^j \qquad \text{(i.e., observation sequences of length no more than the finite horizon)}$$

$$R'(s, \langle a_1, \dots, a_m \rangle, \boldsymbol{\sigma}) = R(s, a_1, \dots, a_m)$$

The DEC-POMDP assumes perfect recall, so we use the state estimator functions from Equations 5 and 6. Since there is no communication for this COM-MTDP, we have a fixed silent policy, $\boldsymbol{\pi}_{\alpha\Sigma}$. We can translate any domain-level policy, $\boldsymbol{\pi}_{\alpha A}$, into a DEC-POMDP joint policy, $\delta$, as follows:

$$\delta^i(o_1^i, \dots, o_t^i) \equiv \pi_{iA}(\langle o_1^i, \dots, o_t^i \rangle) \tag{8}$$

The expected utility of following this joint policy, $\delta$, within the DEC-POMDP is identical to that of following $\boldsymbol{\pi}_{\alpha\Sigma}$ and $\boldsymbol{\pi}_{\alpha A}$ within the constructed COM-MTDP. Thus, there exists a policy with expected utility greater than $K$ for the COM-MTDP if and only if there exists one for the DEC-POMDP. The decision problem for a DEC-POMDP is known to be NEXP-complete, so the COM-MTDP problem must be NEXP-hard.

To show that the COM-MTDP is in NEXP, our proof proceeds similarly to that of the DEC-POMDP. In other words, we guess the joint policy, $\boldsymbol{\pi}_\alpha$, and write it down in exponential time (we assume that $T \leq |S|$). We can take the COM-MTDP plus the policy and generate (in exponential time) a corresponding MDP where the state space is the space of all possible combined belief states of the agents. We can then use dynamic programming to determine (in exponential time) whether $\boldsymbol{\pi}_\alpha$ generates an expected reward of at least $K$. $\square$

In the remainder of this section, we examine the effect of communication on the complexity of constructing team policies that generate optimal behavior. We start by examining the case under the condition of *free communication*, where we would expect the benefit of communication to be the greatest. To begin with, suppose that each agent is capable of communicating its entire observation (i.e., $\Sigma_i \supseteq \Omega_i$). Before we analyze the complexity of the team decision problem, we first prove that the agents should exploit this capability and communicate their true observation, as long as they incur no cost in doing so:

**Theorem 2** *Under* free communication, *consider a team of agents using a communication policy:* $\pi_{i\Sigma}(b_{i\bullet\Sigma}^t) \equiv \Omega_i^t$. *If the domain-level policy $\boldsymbol{\pi}_{\alpha A}$ maximizes $V^T(\boldsymbol{\pi}_{\alpha A}, \boldsymbol{\pi}_{\alpha\Sigma})$, then this combined policy is dominant over any other policies. In other words, for all policies, $\boldsymbol{\pi}_{\alpha A}'$ and $\boldsymbol{\pi}_{\alpha\Sigma}'$, $V^T(\boldsymbol{\pi}_{\alpha A}, \boldsymbol{\pi}_{\alpha\Sigma}) \geq V^T(\boldsymbol{\pi}_{\alpha A}', \boldsymbol{\pi}_{\alpha\Sigma}')$.*

**Proof:** Suppose we have some other communication policy, $\boldsymbol{\pi}_{\alpha\Sigma}'$, that specifies something other than complete communication (e.g., keeping quiet, lying). Suppose that there is some domain-level policy, $\boldsymbol{\pi}_{\alpha A}'$, that allows the team to attain some expected reward, $K$, when used in combination with $\boldsymbol{\pi}_{\alpha\Sigma}'$. Then, we can construct a domain-level policy, $\boldsymbol{\pi}_{\alpha A}$, such that the team attains the same expected reward, $K$, when used in conjunction with the complete-communication policy, $\boldsymbol{\pi}_{\alpha\Sigma}$, as defined in the statement of Theorem 2.

The communication policy, $\boldsymbol{\pi}_{\alpha\Sigma}'$, produces a different set of belief states (denoted $b_{i\bullet\Sigma}'^t$ and $b_{i\Sigma\bullet}'^t$) than those for $\boldsymbol{\pi}_{\alpha\Sigma}$ (denoted $b_{i\bullet\Sigma}^t$ and $b_{i\Sigma\bullet}^t$). In particular, we use state estimator functions, $SE'_{i\bullet\Sigma}$ and $SE'_{i\Sigma\bullet}$ as defined in Equations 5 and 6 to generate $b'^t_{i\bullet\Sigma}$ and $b'^t_{i\Sigma\bullet}$. Each belief state is a complete history of observation and communication pairs for each agent. On the other hand, under the complete communication of $\boldsymbol{\pi}_{\alpha\Sigma}$, the state estimator functions of Equations 5 and 6 reduce to:

$$SE_{i\bullet\Sigma}(\langle \boldsymbol{\Omega}^0_\alpha, \dots, \boldsymbol{\Omega}^{t-1}_\alpha \rangle, \Omega^t_i) = \langle \boldsymbol{\Omega}^0_\alpha, \dots, \boldsymbol{\Omega}^{t-1}_\alpha, \Omega^t_i \rangle \tag{9}$$

$$SE_{i\Sigma\bullet}(\langle \boldsymbol{\Omega}^0_\alpha, \dots, \boldsymbol{\Omega}^{t-1}_\alpha, \Omega^t_i \rangle, \boldsymbol{\Sigma}^t_\alpha) = \langle \boldsymbol{\Omega}^0_\alpha, \dots, \boldsymbol{\Omega}^{t-1}_\alpha, \boldsymbol{\Sigma}^t_\alpha \rangle$$
$$= \langle \boldsymbol{\Omega}^0_\alpha, \dots, \boldsymbol{\Omega}^{t-1}_\alpha, \boldsymbol{\Omega}^t_\alpha \rangle \tag{10}$$

Thus, $\boldsymbol{\pi}_{\alpha A}$ is defined over a different set of belief states than $\boldsymbol{\pi}'_{\alpha A}$. In order to determine an equivalent $\boldsymbol{\pi}_{\alpha A}$, we must first define a recursive mapping, $\boldsymbol{m}$, that translates the belief states defined by $\boldsymbol{\pi}_{\alpha\Sigma}$ into those defined by $\boldsymbol{\pi}'_{\alpha\Sigma}$:

$$m_i(b^t_{i\Sigma\bullet}) = m_i\left(\langle b^{t-1}_{i\Sigma\bullet}, \boldsymbol{\Omega}^t_\alpha \rangle\right) = m_i\left(\langle b^{t-1}_{i\Sigma\bullet}, \langle \Omega^t_i, \boldsymbol{\Omega}^t_\alpha \rangle \rangle\right)$$

$$= \left\langle m_i(b^{t-1}_{i\Sigma\bullet}), \langle \Omega^t_i, \boldsymbol{\Sigma}'^t_\alpha \rangle \right\rangle = \left\langle m_i(b^{t-1}_{i\Sigma\bullet}), \left\langle \Omega^t_i, \prod_{j\in\alpha} \Sigma'^t_j \right\rangle \right\rangle$$

$$= \left\langle m_i(b^{t-1}_{i\Sigma\bullet}), \left\langle \Omega^t_i, \prod_{j\in\alpha} \pi'_{j\Sigma}(SE'_{j\bullet\Sigma}(m_j(b^{t-1}_{j\Sigma\bullet}), \Omega^t_j)) \right\rangle \right\rangle \tag{11}$$

Given this mapping, we then specify: $\pi_{iA}(b^t_{i\Sigma\bullet}) = \pi'_{iA}(m_i(b^t_{i\Sigma\bullet}))$. Executing this domain-level policy, in conjunction with the communication policy, $\boldsymbol{\pi}_{\alpha\Sigma}$, results in the identical behavior as execution of the alternate policies, $\boldsymbol{\pi}'_{\alpha A}$ and $\boldsymbol{\pi}'_{\alpha\Sigma}$. Therefore, the team following the policies, $\boldsymbol{\pi}_{\alpha A}$ and $\boldsymbol{\pi}_{\alpha\Sigma}$ will achieve the same expected value of $K$, as under $\boldsymbol{\pi}'_{\alpha A}$ and $\boldsymbol{\pi}'_{\alpha\Sigma}$. $\square$

Given this dominance of the complete-communication policy, we can prove that the problem of constructing teams that coordinate optimally is simpler when communication is free.

**Theorem 3** *The decision problem of determining whether there exist policies, $\boldsymbol{\pi}_{\alpha\Sigma}$ and $\boldsymbol{\pi}_{\alpha A}$, for a given COM-MTDP with free communication under collective partial observability, that yield a total reward at least $K$ over some finite horizon $T$ is PSPACE-complete.*

**Proof:** To prove that the problem is PSPACE-hard, we reduce the single-agent POMDP to a COM-MTDP. In particular, if we are given a POMDP, $\langle S, A, P, \Omega, O, R \rangle$, we can construct a COM-MTDP, $\langle S', A'_1, \Sigma'_1, P', \Omega'_1, O'_1, B'_1, R' \rangle$, for a single-agent team (i.e., $\alpha = \{1\}$):

$S' = S$

$A'_1 = A$

$\Sigma'_1 = \emptyset$

$P'(s, \langle a_1 \rangle, s') = P(s, a_1, s')$

$\Omega'_1 = \Omega$

$O_1'(s, \langle a_1 \rangle, \langle \omega_1 \rangle) = O(s, a_1, \omega_1)$

$B_1' = \cup_{j=1}^{T} (\Omega)^j$      (i.e., observation sequences of length no more than the finite horizon)

$R_A'(s, \langle a_1 \rangle) = R(s, a_1)$

$R_\Sigma'(s, \boldsymbol{\sigma}) = 0$

This COM-MTDP satisfies our assumption of free communication. The POMDP assumes perfect recall, so we use the state estimator functions from Equations 5 and 6. Just as in the proof of Theorem 1, we can show that there exists a policy with expected utility greater than $K$ for this COM-MTDP if and only if there exists one for the POMDP. The decision problem for the POMDP is known to be PSPACE-hard (Papadimitriou & Tsitsiklis, 1987), so the COM-MTDP problem under free communication must be PSPACE-hard.

To show that the problem is in PSPACE, we take a COM-MTDP under free communication and reduce it to a single-agent POMDP. In particular, if we are given a COM-MTDP, $\langle S, \boldsymbol{A}_\alpha, \boldsymbol{\Sigma}_\alpha, P, \boldsymbol{\Omega}_\alpha, \boldsymbol{O}_\alpha, \boldsymbol{B}_\alpha, R \rangle$, we can construct a single-agent POMDP, $\langle S', A', P', \Omega', O', R' \rangle$, as follows:

$S' = S$

$A' = \boldsymbol{A}_\alpha$

$P'(s, \boldsymbol{a}, s') = P(s, \boldsymbol{a}, s')$

$\Omega' = \boldsymbol{\Omega}_\alpha$

$O'(s, \boldsymbol{a}, \boldsymbol{\omega}) = \boldsymbol{O}_\alpha(s, \boldsymbol{a}, \boldsymbol{\omega})$

$R'(s, \boldsymbol{a}) = R_A(s, \boldsymbol{a})$

From Theorem 2, we need to consider only the complete-communication policy for the COM-MTDP and this policy has a zero reward. Therefore, the decision problem for the COM-MTDP is simply to find a domain-level policy that produces an expected reward exceeding $K$. Given full communication, the state estimator functions for the COM-MTDP (as shown in the proof of Theorem 2) reduce to Equation 10. A policy for our POMDP specifies an action for each and every history of observations: $\pi' : \cup_{j=1}^{T} (\Omega')^j \to A'$. The history of observations for the single-agent POMDP corresponds to the belief states of our COM-MTDP under full communication. Therefore, we can translate a POMDP-policy, $\pi'$, into an equivalent domain-level policy for the COM-MTDP:

$$\pi_A(\langle \boldsymbol{\omega}_0, \boldsymbol{\omega}_1, \ldots, \boldsymbol{\omega}_t \rangle) \equiv \pi'(\langle \boldsymbol{\omega}_0, \boldsymbol{\omega}_1, \ldots, \boldsymbol{\omega}_t \rangle) \tag{12}$$

A team following $\pi_A$ will perform the exact same domain-level actions as a single agent following $\pi'$. Thus, there exists a policy with expected utility greater than $K$ for the COM-MTDP if and only if there exists one for the POMDP. The decision problem for a POMDP is known to be in PSPACE (Papadimitriou & Tsitsiklis, 1987), so the COM-MTDP problem (under free communication) must be in PSPACE as well. □

**Theorem 4** *The decision problem of determining whether there exist policies,* $\pi_{\alpha\Sigma}$ *and* $\pi_{\alpha A}$*, for a given COM-MTDP with* free communication *and* collective observability*, that yield a total reward at least K over some finite horizon T is P-complete.*

**Proof:** The proof follows that of Theorem 3, but with a reduction to and from the MDP decision problem, rather than the POMDP. The MDP decision problem is P-complete (Papadimitriou & Tsitsiklis, 1987). □

**Theorem 5** *The decision problem of determining whether there exist policies,* $\pi_{\alpha\Sigma}$ *and* $\pi_{\alpha A}$*, for a given COM-MTDP with* individual observability*, that yield a total reward at least K over some finite horizon T (given integers K and T) is P-complete.*

**Proof:** The proof follows that of Theorem 4, except that we can reduce the problem to and from an MDP regardless of what communication policy the team uses. □

**Theorem 6** *The decision problem of determining whether there exist policies,* $\pi_{\alpha\Sigma}$ *and* $\pi_{\alpha A}$*, for a given COM-MTDP with* non-observability*, that yield a total reward at least K over some finite horizon T (given integers K and T) is NP-complete.*

**Proof:** The proof follows that of Theorem 4, except that we can reduce the problem to and from an single-agent non-observable MDP (NOMDP) regardless of what communication policy the team uses. In particular, because the agents are all equally ignorant of the state, communication has no effect. The NOMDP decision problem is NP-complete (Papadimitriou & Tsitsiklis, 1987). □

Thus, we have used the COM-MTDP framework to characterize the difficulty of problem domains in agent teamwork along the dimensions of communication cost and observability. Table 2 summarizes our results, which we can use in deciding where to concentrate our energies in attacking teamwork problems. We can use these results to draw some conclusions about the challenges to designers of multiagent teams:

- The greatest challenges lie in those domains with either *collective observability* or *collective partial observability* and with nonzero communication cost.

- Under *collective observability* and *collective partial observability*, teamwork without communication is highly intractable, but, with *free communication*, the complexity becomes on par with that of single-agent planning problems.

- Agent team designers have much to gain by increasing the observational capabilities of their team (e.g., by adding new sensor agents) because of the reduction in complexity gained by making the domain *collectively observable*.

- Furthermore, the results from Theorems 3 and 4 hold in any domain where the result from Theorem 2 holds (i.e., when complete communication is the dominant policy). Therefore, while perfectly free communication may be rare, these results show that investment in communication in teamwork can pay off with a significant simplification of optimal teamwork.

|  | Individually Observable | Collectively Observable | Collectively Partially Observable | Non-Observable |
|---|---|---|---|---|
| No Comm. | P-complete | NEXP-complete | NEXP-complete | NP-Complete |
| General Comm. | P-complete | NEXP-complete | NEXP-complete | NP-Complete |
| Free Comm. | P-complete | P-complete | PSPACE-complete | NP-Complete |

Table 2: Time complexity of COM-MTDPs.

- On the other hand, when the world is *individually observable* or *non-observable*, communication makes no difference in performance.

- It should be noted that even under those conditions where the problem is P-complete, the complexity of optimal teamwork is polynomial in the number of states of the world, which may still be impractically high.

- The above complexity results pertain to finding policies that are optimal *subject to the domain properties*. We will find different expected rewards of the optimal policies under different observability and communication properties. For instance, cutting off all of the agents' sensors makes the domain *non-observable* and reduces the complexity of generating an optimal policy from NEXP to NP, but we would expect an associated drop in the expected reward achieved by the team.

## 4. Evaluating Team Coordination

Table 2 shows that providing optimal domain-level and communication policies for teams is a difficult challenge. Many systems alleviate this difficulty by having domain experts provide the domain-level plans (Tambe, 1997; Tidhar, 1993). Then, the problem for the agents reduces to generating the appropriate team coordination, $\pi_{\alpha\Sigma}$, to ensure that they properly execute the domain-level plans, $\pi_{\alpha A}$. In this section, we demonstrate the COM-MTDP framework's ability to analyze existing teamwork approaches in the literature. Our methodology for such analysis begins by encoding such a teamwork method as a communication-level policy. In other words, we translate the method into an algorithm that maps agent beliefs (e.g., observation sequences) into communication decisions. To evaluate the performance of this policy, we then instantiate a COM-MTDP that represents the states, transition probabilities, and reward function of a domain of interest. Our methodology provides an evaluation of the policy in terms of the expected reward earned by the team when following the policy in the specified domain.

We demonstrate this methodology by using our COM-MTDP framework to analyze joint intentions theory (Cohen & Levesque, 1991b, 1991a; Levesque et al., 1990), which provides a common basis for many existing approaches to team coordination. Section 4.1 models two key instantiations of joint intentions taken from the literature (Jennings, 1995; Tambe, 1997) as COM-MTDP communication policies. Section 4.2 analyzes the conditions under which these policies generate optimal behavior and provides a third candidate policy that makes communication decisions that are locally optimal within the context of joint intentions. In addition to providing the results for the particular team coordination strategies investigated, this section also illustrates a general methodology by which one can use our COM-MTDP framework to encode and evaluate coordination strategies proposed by existing multiagent research.

## 4.1 Joint Intentions in a COM-MTDP

Joint intention theory provides a prescriptive framework for multiagent coordination in a team setting. It does not make any claims of optimality in its teamwork, but it provides theoretical justifications for its prescriptions, grounded in the attainment of mutual belief among the team members. We can use the COM-MTDP framework to identify the domain properties under which attaining mutual belief generates optimal behavior and to quantify precisely how suboptimal the performance will be otherwise.

Joint intentions theory requires that team members jointly commit to a joint persistent goal, $G$. It also requires that when any team member privately believes that $G$ is achieved (or unachievable or irrelevant), it must then attain mutual belief throughout the team about this achievement (or unachievability or irrelevance). To encode this prescription of joint intentions theory within our COM-MTDP model, we first specify the joint goal, $G$, as a subset of states, $G \subseteq S$, where the desired goal is achieved (or unachievable or irrelevant).

Presumably, such a prescription indicates that joint intentions are not specifically intended for *individually observable* environments. Upon achieving the goal in an *individually observable* environment, each agent would simultaneously observe that $S^t \in G$. Because of our assumption that the COM-MTDP model components (including $\mathbf{O}_\alpha$) are common knowledge to the team, each agent would also simultaneously come to believe that its teammates have observed that $S^t \in G$, and that its teammates believe that it believes that all of the team members have observed that $S^t \in G$, and so on. Thus, the team immediately attains mutual belief in the achievement of the goal under *individual observability* without any additional communication necessary by the team.

Instead, the joint intention framework aims at domains with some degree of unobservability. In such domains, the agents must signal the other agents, either through communication or some informative domain-level action, to attain mutual belief. However, we can also assume that joint intention theory does not focus on domains with *free communication*, where Theorem 2 shows that we can simply have the agents communicate everything, all the time, without the need for more complex prescriptions.

The joint intention framework does not specify a precise communication policy for the attainment of mutual belief. In this paper, we focus on communication only in the case of goal achievement, but our methodology extends to handle unachievability and irrelevance as well. One well-known approach (Jennings, 1995) applied joint intentions theory by having the agents communicate the achievement of the joint goal, $G$, as soon as they believe $G$ to be true. To instantiate the behavior of Jennings' agents within a COM-MTDP, we construct a communication policy, $\boldsymbol{\pi}_{\alpha\Sigma}^J$, that specifies that an agent sends the special message, $\sigma_G$, when it first believes that $G$ holds. Following joint intentions' assumption of *sincerity* (Smith & Cohen, 1996), we require that the agents never select the special $\sigma_G$ message in a belief state unless they believe $G$ to be true with certainty. With this requirement and with our assumption of the team's common knowledge of the communication model, we can assume that all of the other agents immediately accept the special message, $\sigma_G$, as true, and that the agents know that all their team members accept the message as true, and so on. Thus, the team attains mutual belief that $G$ is true immediately upon receiving the message, $\sigma_G$. We can construct the communication policy, $\boldsymbol{\pi}^J_{\alpha\Sigma}$, in constant time.

The STEAM algorithm is another instantiation of joint intentions that has had success in several real-world domains (Tambe, 1997; Pynadath et al., 1999; Tambe, Pynadath, Chauvat, Das, & Kaminka, 2000; Pynadath & Tambe, 2002). Unlike Jennings' instantiation, the STEAM teamwork model includes decision-theoretic communication selectivity. A domain specification includes two parameters for each joint commitment, $G$: $\tau$, the probability of miscoordinated termination of $G$; and $C_{mt}$, the cost of miscoordinated termination of $G$. In this context, "miscoordinated termination" means that some agents immediately observe that the team has achieved $G$ while the rest do not. STEAM's domain specification also includes a third parameter, $C_c$, to represent the cost of communication of a fact (e.g., the achievement of $G$). Using these parameters, the STEAM algorithm evaluates whether the expected cost of miscoordination outweighs the cost of communication. STEAM expresses this criterion as the following inequality: $\tau \cdot C_{mt} > C_c$. We can define a communication policy, $\boldsymbol{\pi}^S_{\alpha\Sigma}$ based on this criterion: if the inequality holds, then an agent that has observed the achievement of $G$ will send the message, $\sigma_G$; otherwise, it will not. We can construct $\boldsymbol{\pi}^S_{\alpha\Sigma}$ in constant time.

## 4.2 Locally Optimal Policy

Although the STEAM policy is more selective than Jennings', it remains unanswered whether it is *optimally* selective, and researchers continue to struggle with the question of when agents should communicate (Yen et al., 2001). The few reports of suboptimal (in particular, excessive) communication in STEAM characterized the phenomenon as an exceptional circumstance, but it is also possible that STEAM's optimal performance is the exception. We use the COM-MTDP model to derive an analytical characterization of optimal communication here, while Section 5 provides an empirical one by creating an algorithm using that characterization.

Both policies, $\boldsymbol{\pi}^J_{\alpha\Sigma}$, and $\boldsymbol{\pi}^S_{\alpha\Sigma}$ consider sending $\sigma_G$ only when an agent first believes that $G$ has been achieved. Once an agent has the relevant belief, they make different choices, and we consider here what the optimal decision is at this point. The domain is not individually observable, so certain agents may be unaware of the achievement of $G$. When not sending the $\sigma_G$ message, these unaware agents may unnecessarily continue performing actions in the pursuit of achieving $G$. The performance of these extraneous actions could potentially incur costs and lead to a lower utility than one would expect when sending the $\sigma_G$ message.

The decision to send $\sigma_G$ or not matters only if the team achieves $G$ *and* one agent comes to know this fact. We define the random variable, $T_G$, to be the earliest time at which an agent knows this fact. We denote agent $K_G$ as the agent who knows of the achievement at time $T_G$. If $K_G = i$, for some agent, $i$, and $T_G = t_0$, then agent $i$ has some pre-communication belief state, $b^{t_0}_{i\bullet\Sigma} = \beta$, that indicates that $G$ has been achieved. To more precisely quantify the difference between agent $i$ sending the $\sigma_G$ message at time $T_G$ vs.

never sending it, we define the following value:

$$\Delta^T(t_0, i, \beta) \equiv E\left[\sum_{t=0}^{T-t_0} R^{t_0+t}\,\middle|\,\Sigma_i^{t_0} = \sigma_G, T_G = t_0, K_G = i, b_{i\bullet\Sigma}^{t_0} = \beta\right]$$

$$- E\left[\sum_{t=0}^{T-t_0} R^{t_0+t}\,\middle|\,\Sigma_i^{t_0} = \text{null}, T_G = t_0, K_G = i, b_{i\bullet\Sigma}^{t_0} = \beta\right] \quad (13)$$

We assume that, for all times other than $T_G$, the agents follow some communication policy, $\boldsymbol{\pi}_{\alpha\Sigma}$, that never specifies $\sigma_G$. Thus, $\Delta^T$ measures the difference in expected reward that hinges on agent $i$'s specific decision to send or not send $\sigma_G$ at time $t_0$. Given this definition, it is locally optimal for agent $i$ to send the special message, $\sigma_G$, at time $t_0$, if and only if $\Delta^T \geq 0$. We define the communication policy, $\boldsymbol{\pi}_{\alpha\Sigma+\sigma}$, as the communication policy following $\boldsymbol{\pi}_{\alpha\Sigma}$ for all agents at all times, except for agent $i$ under belief state $\beta$, when agent $i$ sends message $\sigma$. With this definition, $\boldsymbol{\pi}_{\alpha\Sigma+\sigma_G}$, is the policy under which agent $i$ communicates the achievement of $G$, and $\boldsymbol{\pi}_{\alpha\Sigma+\text{null}}$ is the policy under which it does not. Therefore, we can alternatively describe agent $i$'s decision criterion as choosing $\boldsymbol{\pi}_{\alpha\Sigma+\sigma_G}$ over $\boldsymbol{\pi}_{\alpha\Sigma+\text{null}}$ if and only if $\Delta^T \geq 0$.

Unfortunately, while Equation 13 identifies an exact criterion for locally optimal communication, this criterion is not yet operational. In other words, we can not directly implement it as a communication policy for the agents. Furthermore, Equation 13 hides the underlying complexity of the computation involved, which is one of the key goals of our analysis. Therefore, we use the COM-MTDP model to derive an operational expression of $\Delta^T \geq 0$. For simplicity, we define notational shorthand for various sequences and combinations of values. We define a partial sequence of random variables, $X^{<t}$, to be the sequence of random variables for all times before it: $X^0, X^1, \dots, X^{t-1}$. We make similar definitions for the other relational operators (i.e., $X^{>t}$, $X^{\geq t}$, etc.). The expression, $(S)^T$, denotes the cross product over states of the world, $\prod_{t=0}^{T} S$, as distinguished from the time-indexed random variable, $S^T$, which denotes the value of the state at time $T$. The notation, $s^{\geq t_0}[t]$, specifies the element in slot $t$ within the vector $s^{\geq t_0}$. We define the function, $\Upsilon$, as shorthand within our probability expressions. It allows us to compactly represent a particular subsequence of world and agent belief states occurring, conditioned on the current situation, as follows:

$$\Pr\left(\Upsilon\left(\langle t, t'\rangle, s, \boldsymbol{\beta}_{\bullet\Sigma}\right)\right) \equiv \Pr(S^{\geq t, \leq t'} = s, \boldsymbol{b}_{\alpha\bullet\Sigma}^{\geq t, \leq t'} = \boldsymbol{\beta}_{\bullet\Sigma}\,\middle|\,T_G = t_0, K_G = i, b_{i\bullet\Sigma}^{t_0} = \beta) \quad (14)$$

Informally, $\Upsilon\left(\langle t, t'\rangle, s, \boldsymbol{\beta}_{\bullet\Sigma}\right)$ represents the event that the world and belief states from time $t$ through $t'$ correspond to the specified sequences, $s$ and $\boldsymbol{\beta}_{\bullet\Sigma}$, respectively, conditioned on agent $i$ being the first to know of $G$'s achievement at time $t_0$ with a belief state, $\beta$. We define the function, $\beta_{\Sigma\bullet}$, to map a pre-communication belief state into the post-communication belief state that arises from a communication policy:

$$\beta_{\Sigma\bullet}(\boldsymbol{\beta}_{\bullet\Sigma}, \boldsymbol{\pi}_{\alpha\Sigma}) \equiv \boldsymbol{SE}_{\alpha\Sigma\bullet}(\boldsymbol{\beta}_{\bullet\Sigma}, \boldsymbol{\pi}_{\alpha\Sigma}(\boldsymbol{\beta}_{\bullet\Sigma})) \quad (15)$$

This definition of $\beta_{\Sigma\bullet}$ is a well-defined function because of the deterministic nature of the policy, $\boldsymbol{\pi}_{\alpha\Sigma}$, and state-estimator function, $\boldsymbol{SE}_{\alpha\Sigma\bullet}$.

**Theorem 7** *If we assume that, upon achievement of $G$, no communication other than $\sigma_G$ is possible, then the condition $\Delta^T(t_0, i, \beta) \geq 0$ holds if and only if:*

$$
\sum_{s^{\leq t_0} \in (S)^{t_0}} \sum_{\boldsymbol{\beta}_{\bullet\Sigma}^{\leq t_0} \in \boldsymbol{B}_\alpha^{t_0}} \Pr(\Upsilon(\langle 0, t_0 \rangle, s^{\leq t_0}, \beta_{\bullet\Sigma}^{\leq t_0}))
$$

$$
\cdot \left( \sum_{s^{\geq t_0} \in (S)^{T-t_0+1}} \sum_{\boldsymbol{\beta}_{\bullet\Sigma}^{\geq t_0} \in \boldsymbol{B}_\alpha^{T-t_0+1}} \Pr\left( \Upsilon(\langle t_0, T\rangle, s^{\geq t_0}, \beta_{\bullet\Sigma}^{\geq t_0}) \,\Big|\, \Sigma_i^{t_0} = \sigma_G, \Upsilon(\langle 0, t_0\rangle, s^{\leq t_0}, \beta_{\bullet\Sigma}^{\leq t_0}) \right) \right.
$$

$$
\cdot \sum_{t=t_0}^{T} R_A\left( s^{\geq t_0}[t], \pi_{\alpha A}\left( \boldsymbol{\beta}_{\Sigma\bullet}\left( \boldsymbol{\beta}_{\bullet\Sigma}^{\geq t_0}[t], \pi_{\alpha\Sigma+\sigma_G} \right) \right) \right)
$$

$$
- \sum_{s^{\geq t_0} \in (S)^{T-t_0+1}} \sum_{\boldsymbol{\beta}_{\bullet\Sigma}^{\geq t_0} \in \boldsymbol{B}_\alpha^{T-t_0+1}} \Pr\left( \Upsilon(\langle t_0, T\rangle, s^{\geq t_0}, \beta_{\bullet\Sigma}^{\geq t_0}) \,\Big|\, \Sigma_i^{t_0} = null, \Upsilon(\langle 0, t_0\rangle, s^{\leq t_0}, \beta_{\bullet\Sigma}^{\leq t_0}) \right)
$$

$$
\left. \cdot \sum_{t=t_0}^{T} R_A\left( s^{\geq t_0}[t], \pi_{\alpha A}\left( \boldsymbol{\beta}_{\Sigma\bullet}\left( \boldsymbol{\beta}_{\bullet\Sigma}^{\geq t_0}[t], \pi_{\alpha\Sigma+null} \right) \right) \right) \right)
$$

$$
\geq - \sum_{s \in G} \sum_{\boldsymbol{\beta} \in \boldsymbol{B}_\alpha} \Pr\left( \Upsilon(\langle t_0, t_0\rangle, s, \boldsymbol{\beta}) \right) R_\Sigma(s, \sigma_G) \tag{16}
$$

**Proof:** The complete proof of the following theorem appears in Online Appendix 1. The definition of $\Delta^T$ in Equation 13 is the difference between two expectations, where each expectation is a sum over the possible trajectories of the agent team. Each trajectory must includes a sequence of possible world states, since the agents' reward at each point in time depends on the particular state of the world at that time. The agents' reward also depends on their actions (both domain- and communication-level). These actions are deterministic, given the agents' policies, $\boldsymbol{\pi}_{\alpha A}$ and $\boldsymbol{\pi}_\Sigma$, and their belief states. Thus, in addition to summing over the possible states of the world, we must also sum over the possible states of the agents' beliefs (both pre- and post-communication):

$$
\begin{aligned}
&\Delta^T(t_0, i, \beta) \\
&= \sum_{s^{\leq T} \in (S)^T} \sum_{\boldsymbol{\beta_{\bullet \Sigma}}^{\leq T} \in (\mathbf{B})^T} \sum_{\boldsymbol{\beta_{\Sigma \bullet}}^{\leq T} \in (\mathbf{B})^T} \mathrm{Pr}\left(S^{\leq T} = s^{\leq T}, \mathbf{b}_{\bullet \Sigma}^{\leq T} = \boldsymbol{\beta_{\bullet \Sigma}}^{\leq T}, \mathbf{b}_{\Sigma \bullet}^{\leq T} = \boldsymbol{\beta_{\Sigma \bullet}}^{\leq T}\right. \\
&\hspace{8cm} \left.| \Sigma_i^{t_0} = \sigma_G, T_G = t_0, K_G = i, b_{i\bullet \Sigma}^{t_0} = \beta\right) \\[2mm]
&\hspace{4cm} \cdot \sum_{t=0}^{T} R(s^{\leq T}[t], \boldsymbol{\pi}_A(\boldsymbol{\beta_{\Sigma \bullet}}^{\leq T}[t]), \boldsymbol{\pi}_\Sigma(\boldsymbol{\beta_{\bullet \Sigma}}^{\leq T}[t])) \\
&- \sum_{s^{\leq T} \in (S)^T} \sum_{\boldsymbol{\beta_{\bullet \Sigma}}^{\leq T} \in (\mathbf{B})^T} \sum_{\boldsymbol{\beta_{\Sigma \bullet}}^{\leq T} \in (\mathbf{B})^T} \mathrm{Pr}\left(S^{\leq T} = s^{\leq T}, \mathbf{b}_{\bullet \Sigma}^{\leq T} = \boldsymbol{\beta_{\bullet \Sigma}}^{\leq T}, \mathbf{b}_{\Sigma \bullet}^{\leq T} = \boldsymbol{\beta_{\Sigma \bullet}}^{\leq T}\right. \\
&\hspace{8cm} \left.| \Sigma_i^{t_0} = \mathrm{null}, T_G = t_0, K_G = i, b_{i\bullet \Sigma}^{t_0} = \beta\right) \\[2mm]
&\hspace{4cm} \cdot \sum_{t=0}^{T} R(s^{\leq T}[t], \boldsymbol{\pi}_A(\boldsymbol{\beta_{\Sigma \bullet}}^{\leq T}[t]), \boldsymbol{\pi}_\Sigma(\boldsymbol{\beta_{\bullet \Sigma}}^{\leq T}[t])) \quad (17)
\end{aligned}
$$

We can rewrite these summations more simply using our various shorthand notations:

$$
\begin{aligned}
&= \sum_{s^{\leq T} \in (S)^T} \sum_{\boldsymbol{\beta_{\bullet \Sigma}}^{\leq T} \in (\mathbf{B})^T} \mathrm{Pr}(\Upsilon(\langle 0, T \rangle, s, \boldsymbol{\beta_{\bullet \Sigma}}^{\leq T}) | \Sigma_i^{t_0} = \sigma_G) \\
&\hspace{3cm} \cdot \sum_{t=0}^{T} R(s^{\leq T}[t], \boldsymbol{\pi}_A(\beta_{\Sigma \bullet}(\boldsymbol{\beta_{\bullet \Sigma}}^{\leq T}[t], \boldsymbol{\pi}_{\Sigma \sigma_G})), \boldsymbol{\pi}_{\Sigma \sigma_G}(\boldsymbol{\beta_{\bullet \Sigma}}^{\leq T}[t])) \\
&- \sum_{s^{\leq T} \in (S)^T} \sum_{\boldsymbol{\beta_{\bullet \Sigma}}^{\leq T} \in (\mathbf{B})^T} \mathrm{Pr}(\Upsilon(\langle 0, T \rangle, s, \boldsymbol{\beta_{\bullet \Sigma}}^{\leq T}) | \Sigma_i^{t_0} = \mathrm{null}) \\
&\hspace{3cm} \cdot \sum_{t=0}^{T} R(s^{\leq T}[t], \boldsymbol{\pi}_A(\beta_{\Sigma \bullet}(\boldsymbol{\beta_{\bullet \Sigma}}^{\leq T}[t], \boldsymbol{\pi}_{\Sigma \mathrm{null}})), \boldsymbol{\pi}_{\Sigma \mathrm{null}}(\boldsymbol{\beta_{\bullet \Sigma}}^{\leq T}[t])) \quad (18)
\end{aligned}
$$

The remaining derivation exploits our Markovian assumptions to rearrange the summations and cancel like terms to produce the theorem's result. □

Theorem 7 states, informally, that we prefer sending $\sigma_G$ whenever the the cost of execution after achieving $G$ outweighs the cost of communication of the fact that $G$ has been achieved. More precisely, the outer summations on the left-hand side of the inequality iterate over all possible past histories of world and belief states, producing a probability distribution over the possible states the team can be in at time $t_0$. For each such state, the expression inside the parentheses computes the difference in domain-level reward, over all possible *future* sequences of world and belief states, between sending and not sending $\sigma_G$. By our theorem's assumption that no communication other than $\sigma_G$ is possible after $G$ has been achieved, we can ignore any communication costs in the future. However, if we relax this assumption, we can extend the left-hand side in a straightforward manner into a longer expression that accounts for the difference in future communication costs as well. Thus, the left-hand side captures our intuition that, when not communicating, the team will incur a cost if the agents other than $i$ are unaware of $G$'s achievement. The right-hand side of the inequality is a summation of the cost of sending the $\sigma_G$ message over possible current states and belief states.

|  | Individually Observable | Collectively Observable | Collectively Partially Observable | Non-Observable |
|---|---|---|---|---|
| No Comm. | $\Omega(1)$ | $\Omega(1)$ | $\Omega(1)$ | $\Omega(1)$ |
| General Comm. | $\Omega(1)$ | $O((|S| \cdot |\mathbf{\Omega}_\alpha|)^T)$ | $O((|S| \cdot |\mathbf{\Omega}_\alpha|)^T)$ | $\Omega(1)$ |
| Free Comm. | $\Omega(1)$ | $\Omega(1)$ | $\Omega(1)$ | $\Omega(1)$ |

Table 3: Time complexity of locally optimal decision.

We can use Theorem 7 to derive the locally optimal communication decision across various classes of problem domains. Under *no communication*, we cannot send $\sigma_G$. Under *free communication*, the right-hand side is 0, so the inequality is always true, and we know to prefer sending $\sigma_G$. Under no assumptions about communication, the determination is more complicated. When the domain is *individually observable*, the left-hand side becomes 0, because *all* of the agents know that $G$ has been achieved (and thus there is no difference in execution when sending $\sigma_G$). Therefore, the inequality is always false (unless under *free communication*), and we prefer not sending $\sigma_G$. When the environment is not individually observable and communication is available but not free, then, to be locally optimal at time $t_0$, agent $i$ must evaluate Inequality 16 in its full complexity. Since the inequality sums rewards over all possible sequences of states and observations, the time complexity of the corresponding algorithm is $O((|S| \cdot |\mathbf{\Omega}_\alpha|)^T)$. While this complexity is unacceptable for most real-world problems, it still provides an exponential savings over searching the entire policy space for the globally optimal policy, where any agent could potentially send $\sigma_G$ at times other than $T_G$. Table 3 provides a table of the complexity required to determine the locally optimal policy under the various domain properties.

We can now show that although Theorem 7's algorithm for locally optimal communication provides a significant computational savings over finding the global optimum, it still outperforms existing teamwork models, as exemplified by our $\boldsymbol{\pi}_{\alpha\Sigma}^J$ and $\boldsymbol{\pi}_{\alpha\Sigma}^S$ policies. First, we can use the criterion of Theorem 7 to evaluate the optimality of the policy, $\boldsymbol{\pi}_{\alpha\Sigma}^J$. If $\Delta^T(t_0, i, \beta) \geq 0$ for all possible times $t_0$, agents $i$, and belief states $\beta$ that are consistent with the achievement of the goal $G$, then the locally optimal policy will *always* specify sending $\sigma_G$. In other words, $\boldsymbol{\pi}_{\alpha\Sigma}^J$ will be identical to the locally optimal policy. However, if the inequality of Theorem 7 is *ever* false, then $\boldsymbol{\pi}_{\alpha\Sigma}^J$ is not even locally, let alone globally, optimal.

Second, we can also use Theorem 7 to evaluate STEAM by viewing STEAM's inequality, $\tau \cdot C_{mt} > C_c$, as a crude approximation of Inequality 16. In fact, there is a clear correspondence between the terms in the two inequalities. The left-hand side of Inequality 16 computes an exact expected cost of miscoordination. However, unlike STEAM's monolithic $\tau$ parameter, the optimal criterion evaluates a complete probability distribution over all possible states of miscoordination by considering all possible past sequences consistent with the agent's current beliefs. Likewise, unlike STEAM's monolithic $C_{mt}$ parameter, the optimal criterion looks ahead over all possible future sequences of states to determine the true expected cost of miscoordination. Furthermore, we can view STEAM's parameter, $C_c$, as an approximation of the communication cost computed by the right-hand side of Inequality 16. Again, STEAM uses a single parameter, while the optimal criterion computes an expected cost over all possible states of the world.

STEAM does have *some* flexibility in its representation, because $C_{mt}$, $\tau$, and $C_c$ are not necessarily fixed across the *entire* domain. For instance, $C_{mt}$ may vary based on the specific joint plan that the agents may have jointly committed to (i.e., there may be a different $C_{mt}$ for each goal $G$). Thus, while Theorem 7 suggests significant additional flexibility in computing $C_{mt}$ through explicit lookahead, the optimal criterion derived with the COM-MTDP model also provides a justification for the overall structure behind STEAM's approximate criterion. Furthermore, STEAM's emphasis on on-line computation makes the computational complexity of Inequality 16 (as presented in Table 3) unacceptable, so the approximation error may be acceptable given the gains in efficiency. For a specific domain, we can use empirical evaluation (as demonstrated in the next section) to quantify the error and efficiency to precisely judge this tradeoff.

## 5. Empirical Policy Evaluation

In addition to providing these analytical results over general classes of problem domains, the COM-MTDP framework also supports the analysis of *specific* domains. Given a particular problem domain, we can construct an optimal communication policy or, if the complexity of computing an optimal policy is prohibitive, we can instead evaluate and compare candidate approximate policies. To provide a reusable tool for such evaluations, we have implemented the COM-MTDP model as a Python class with domain-independent methods for the evaluation of arbitrary policies and for the generation of both locally optimal policies using Theorem 7 and globally optimal policies through brute-force search of the policy space. This software is available in Online Appendix 1.

This section presents results of a COM-MTDP analysis of an example domain involving agent-piloted helicopters, where we focus on the key communication decision faced by many multiagent frameworks (as described in Section 4), but vary the cost of communication and degree of observability to generate a space of distinct domains with different implications for the agents' performance. By evaluating communication policies over various configurations of this particular testbed domain, we demonstrate a methodology by which one can use the COM-MTDP framework to model *any* problem domain and to evaluate candidate communication policies for it.

### 5.1 Experimental Setup

Consider two helicopters that must fly across enemy territory to their destination, as illustrated in Figure 1. The first, piloted by agent *Transport*, is a transport vehicle with limited firepower. The second, piloted by agent *Escort*, is an escort vehicle with significant firepower. Somewhere along their path is an enemy radar unit, but its location is unknown (a priori) to the agents. *Escort* is capable of destroying the radar unit upon encountering it. However, *Transport* is not, but it can escape detection by the radar unit by traveling

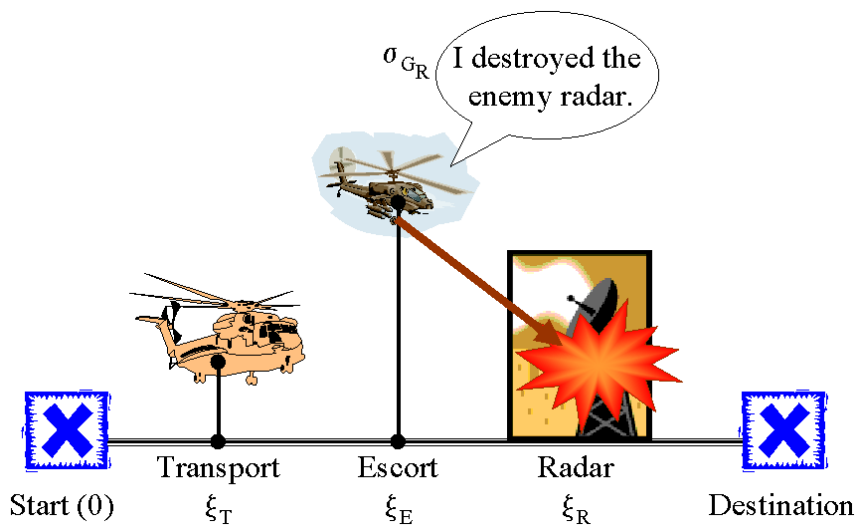

σ_{G_R} "I destroyed the enemy radar."

Start (0)    Transport    Escort    Radar    Destination
             ξ_T          ξ_E       ξ_R

Figure 1: Illustration of helicopter team scenario.

at a very low altitude (*nap-of-the-earth* flight), though at a lower speed than at its typical, higher altitude. In this scenario, *Escort* will not worry about detection, given its superior firepower; therefore, it will fly at a fast speed at its typical altitude.

The two agents form a top-level joint commitment, $G_D$, to reach their destination. There is no incentive for the agents to communicate the achievement of this goal, since they will both eventually reach their destination with certainty. However, in the service of their top-level goal, $G_D$, the two agents also adopt a joint commitment, $G_R$, of destroying the radar unit. We consider here the problem facing *Escort* with respect to communicating the achievement of goal, $G_R$. If *Escort* communicates the achievement of $G_R$, then *Transport* knows that it is safe to fly at its normal altitude (thus reaching the destination sooner). If *Escort* does *not* communicate the achievement of $G_R$, there is still some chance that *Transport* will observe the event anyway. If *Transport* does not observe the achievement of $G_R$, then it must fly nap-of-the-earth the whole distance, and the team receives a lower reward because of the later arrival. Therefore, *Escort* must weigh the increase in expected reward against the cost of communication.

In the COM-MTDP model of this scenario (presented in Figures 2, 3 and 4), the world state is the position (along a straight line between origin and destination) of *Transport*, *Escort*, and the enemy radar. The enemy is at a randomly selected position somewhere in between the agents' initial position and their destination. *Transport* has no possible communication actions, but it can choose between two domain-level actions: flying nap-of-the-earth and flying at its normal speed and altitude. *Escort* has two domain-level actions: flying at its normal speed and destroying the radar. *Escort* also has the option of communicating the special message, $\sigma_{G_R}$, indicating that the radar has been destroyed. In the tables of Figures 2, 3 and 4, the "·" symbol represents a wild-card (or "don't care") entry.

If *Escort* arrives at the radar, then it observes its presence with certainty and can destroy it to achieve $G_R$. The likelihood of *Transport*'s observing the radar's destruction is a function of its distance from the radar. We can vary this function's *observability* parameter

411

| $\alpha$ | $= \{\text{Escort } (E), \text{Transport } (T)\}$ | | | |
|---|---|---|---|---|
| $S$ | $= \Xi_E \times \Xi_T \times \Xi_R$ | | | |
| | **Position of Escort:** $\Xi_E = \{0, 1, \dots, 8, 9, \text{Destination}\}$ | | | |
| | **Position of Transport:** $\Xi_T = \{0, 0.5, \dots, 9, 9.5, \text{Destination},$ | | | |
| | $\text{Destroyed}\}$ | | | |
| | **Position of Radar:** $\Xi_R = \{1, 2, \dots, 8, \text{Destroyed}\}$ | | | |
| $\boldsymbol{A_\alpha}$ | $= A_E \times A_T = \{\text{fly, destroy, wait}\} \times \{\text{fly-NOE, fly-normal, wait}\}$ | | | |
| $\boldsymbol{\Sigma_\alpha}$ | $= \Sigma_E \times \Sigma_T = \{\text{clear } (\sigma_{G_R}), \text{null}\} \times \{\text{null}\}$ | | | |

| | | $\xi_E$ | $\xi_T$ | $\boldsymbol{a}$ | $R_A$ |
|---|---|---|---|---|---|
| $R_A(\langle\xi_E, \xi_T, \xi_R\rangle, \boldsymbol{a})$ | $=$ | $0, \dots, 9$ | $0, \dots, 9.5, \text{Destroyed}$ | $\cdot$ | $0$ |
| | | $0, \dots, 9$ | Destination | $\cdot$ | $r_T$ |
| | | Destination | $0, \dots, 9.5, \text{Destroyed}$ | $\cdot$ | $r_E$ |
| | | Destination | Destination | $\cdot$ | $r_E + r_T$ |

| $R_\Sigma(s, \langle\text{null}, \text{null}\rangle)$ | $= 0$ |
|---|---|
| $R_\Sigma(s, \langle\sigma_{G_R}, \text{null}\rangle)$ | $= -r_\Sigma \in [0, 1]$ |

Figure 2: COM-MTDP model of states, actions, and rewards for helicopter scenario.

($\lambda$ in Figure 4) within the range $[0, 1]$ to generate distinct domain configurations (0 means that *Transport* will never observe the radar's destruction; 1 means *Transport* will always observe it). If the observability is 1, then they achieve mutual belief of the achievement of $G_R$ as soon as it occurs (following the argument presented in Section 4.1). However, for any observability less than 1, there is a chance that the agents will not achieve mutual belief simply by common observation. The helicopters receive a fixed reward for each time step spent at their destination. Thus, for a fixed time horizon, the earlier the helicopters reach there, the greater the team's reward. Since flying nap-of-the-earth is slower than normal speed, *Transport* will switch to its normal flying as soon as it either observes that $G_R$ has been achieved or *Escort* sends the message, $\sigma_{G_R}$. Sending the message is not free, so we impose a variable communication cost ($r_\Sigma$ in Figure 2), also within the range $[0, 1]$.

We constructed COM-MTDP models of this scenario for each combination of observability and communication cost within the range $[0, 1]$ at 0.1 increments. For each combination, we applied the Jennings and STEAM policies, as well as a completely silent policy. For this domain, the policy, $\boldsymbol{\pi}_{\alpha\Sigma}^J$, dictates that *Escort* always communicate $\sigma_{G_R}$ upon destroying the radar. For STEAM, we vary the $\tau$ and $C_c$ parameters with the observability and communication cost parameters, respectively. We used two different settings (*low* and *medium*) for the cost of miscoordination, $C_{mt}$. Following the published STEAM algorithm (Tambe, 1997), *Escort* sends message $\sigma_{G_R}$ if and only if STEAM's inequality $\tau \cdot C_{mt} > C_c$, holds. Thus, the two different settings, *low* and *medium*, for $C_{mt}$ generate two distinct communication policies; the *high* setting is strictly dominated by the other two settings in this domain. We also constructed and evaluated locally and globally optimal policies. In applying each of these policies, we used our COM-MTDP model to compute the expected reward received by the team when following the selected policy. We can uniquely determine this expected reward given the candidate communication policy and the particular observability and communication cost parameters, as well as the COM-MTDP model specified in Figures 2, 3, and 4.

- $P(\langle \xi_{E0}, \xi_{T0}, \xi_{R0} \rangle, \langle a_E, a_T \rangle, \langle \xi_{E1}, \xi_{T1}, \xi_{R1} \rangle) =$
  $P_E(\xi_{E0}, a_E, \xi_{E1}) \cdot P_T(\langle \xi_{T0}, \xi_{R0} \rangle, a_T, \xi_{T1}) \cdot P_R(\langle \xi_{E0}, \xi_{R0} \rangle, a_E, \xi_{R1})$

**Escort:** Initial distribution, $\mathrm{Pr}(\Xi_E^0 = 0) = 1$

| $\xi_{E0}$ | $a_E$ | $\xi_{E1}$ | $P_E$ |
|---|---|---|---|
| Destination | · | Destination | 1 |
| $0, \dots, 8$ | fly | $\xi_{E0} + 1$ | 1 |
| $0, \dots, 8$ | destroy | $\xi_{E0} + 1$ | 1 |
| 9 | fly | Destination | 1 |
| 9 | destroy | Destination | 1 |
| · | wait | $\xi_{E0}$ | 1 |

**Transport:** Initial distribution, $\mathrm{Pr}(\Xi_T^0 = 0) = 1$

| $\xi_{T0}$ | $\xi_{R0}$ | $a_T$ | $\xi_{T1}$ | $P_T$ |
|---|---|---|---|---|
| Destination | · | · | Destination | 1 |
| Destroyed | · | · | Destroyed | 1 |
| $0, \dots, 9$ | · | fly-NOE | $\xi_{T0} + 0.5$ | 1 |
| 9.5 | · | fly-NOE | Destination | 1 |
| $0, \dots, 8.5$ | Destroyed | fly-normal | $\xi_{T0} + 1$ | 1 |
| $9, 9.5$ | Destroyed | fly-normal | Destination | 1 |
| · | $\neq$ Destroyed | fly-normal | Destroyed | 1 |
| · | · | wait | $\xi_{T0}$ | 1 |

**Radar:** Initial distribution, $\forall \xi \in \{1, 2, \dots, 8\}$, $Pr(\Xi_R^0 = \xi) = 0.125$

| $\xi_{E0}$ | $\xi_{R0}$ | $a_E$ | $\xi_{R1}$ | $P_R$ |
|---|---|---|---|---|
| · | $\xi_{E0}$ | destroy | Destroyed | 1 |
| · | · | $\neq$ destroy | $\xi_{R0}$ | 1 |
| · | $\neq \xi_{E0}$ | · | $\xi_{R0}$ | 1 |

Figure 3: COM-MTDP model of transition probabilities for helicopter scenario (excludes zero probability rows).

- $\boldsymbol{\Omega}_\alpha = \Omega_E \times \Omega_T$

  - $\Omega_E = \Xi_E \times \Xi_T \times \Omega_{RE}$, where agent *Escort*'s possible observations of the radar consist of $\Omega_{RE} = \{\text{present}, \text{destroyed}, \text{null}\}$

  - $\Omega_T = \Xi_E \times \Xi_T \times \Omega_{RT}$, where agent *Transport*'s possible observations of the radar consist of $\Omega_{RT} = \{\text{destroyed}, \text{null}\}$

- $\boldsymbol{O}_\alpha(s, \langle a_E, a_T \rangle, \langle \omega_E, \omega_T \rangle) = O_E(s, \langle a_E, a_T \rangle, \omega_E) \cdot O_T(s, \langle a_E, a_T \rangle, \omega_T)$

  - $O_E(\langle \xi_E, \xi_T, \xi_R \rangle, \langle a_E, a_T \rangle, \langle \xi_E, \xi_T, \omega_{RE} \rangle) =$

    | $\xi_E$ | $\xi_R$ | $a_E$ | $\omega_{RE}$ | $O_E$ |
    |---------|---------|-------|---------------|-------|
    | · | destroyed | destroy | destroyed | 1 |
    | · | destroyed | $\neq$ destroy | null | 1 |
    | $\xi_R$ | $1, \dots, 9$ | · | present | 1 |
    | $\neq \xi_R$ | $1, \dots, 9$ | · | null | 1 |

  - $O_T(\langle \xi_E, \xi_T, \xi_R \rangle, \langle a_E, a_T \rangle, \langle \xi_E, \xi_T, \omega_{RT} \rangle) =$

    | $\xi_T$ | $\xi_R$ | $a_E$ | $\omega_{RT}$ | $O_T$ |
    |---------|---------|-------|---------------|-------|
    | $0, \dots, 9.5$ | · | destroy | destroyed | $\lambda e^{-(\xi_R - \xi_T)(1-\lambda)}$ |
    | $0, \dots, 9.5$ | · | destroy | null | $1 - \lambda e^{-(\xi_R - \xi_T)(1-\lambda)}$ |
    | $0, \dots, 9.5$ | · | $\neq$ destroy | null | 1 |
    | destroyed | · | · | null | 1 |

    $\lambda \in [0, 1]$

Figure 4: COM-MTDP model of observability for helicopter scenario. These tables exclude both zero probability rows and input feature columns from which $O$ is independent. For example, both agents' observation functions are independent of the transport's selected action, so neither table includes a $a_T$ column.

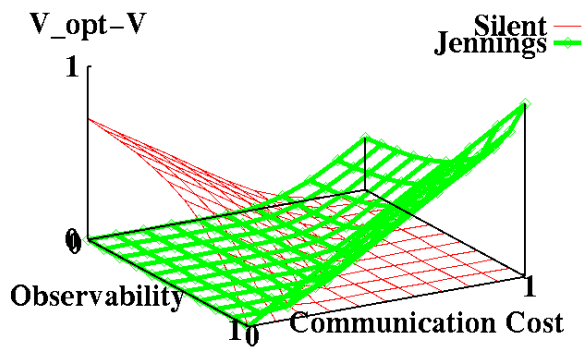

Figure 5: Suboptimality of silent and Jennings policies.

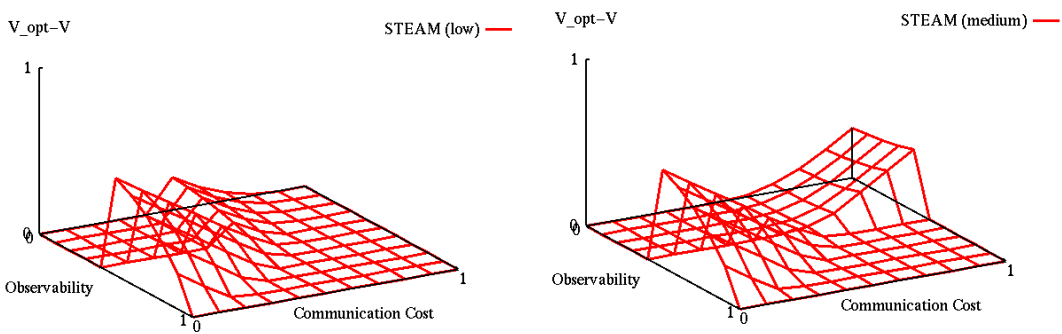

Figure 6: Suboptimality of STEAM policy under both *low* and *medium* costs of miscoordination.

## 5.2 Experimental Results

Figures 5 and 6 plot how much utility the team can expect to lose by following the Jennings, silent, and the two STEAM policies instead of the locally optimal communication policy (thus, higher values mean *worse* performance). We can immediately see that the Jennings and silent policies are significantly suboptimal for many possible domain configurations. For example, not surprisingly, the surface for the policy, $\pi^J_{\alpha\Sigma}$, peaks (i.e., it does most poorly) when the communication cost is high and when the observability is high, while the silent policy does poorly under exactly the opposite conditions.

Previously published results (Jennings, 1995) demonstrated that the Jennings policy led to better team performance by reducing waste of effort produced by alternate policies like our silent one. These earlier results focused on a single domain, and Figure 5 partially confirms their conclusion and shows that the superiority of the Jennings policy over the silent policy extends over a broad range of possible domain configurations. On the other hand, our COM-MTDP results also show that there is a significant subclass of domains (e.g., when communication cost and observability are high) where the Jennings policy is actually *inferior* to the silent policy. Thus, with our COM-MTDP model, we can characterize the types of domains where the Jennings policy outperforms the silent policy and vice versa.

Figure 6 shows the expected value lost by following the two STEAM policies. We can view STEAM as trying to intelligently interpolate between the Jennings and silent policies based on the particular domain properties. In fact, under a *low* setting for $C_{mt}$, we see two thresholds, one along each dimension, at which STEAM switches between following the Jennings and silent policies, and its suboptimality is highest at these thresholds. Under a *medium* setting for $C_{mt}$, STEAM does not exhibit a threshold along the dimension of communication cost, due to the increased cost of miscoordination. Under both settings, STEAM's performance generally follows the better of those two fixed policies, so its maximum suboptimality (0.587 under both settings) is significantly lower than that of the silent (0.700) and Jennings' (1.000) policies. Furthermore, STEAM outperforms the two policies on average, across the space of domain configurations, as evidenced by its mean suboptimality of 0.063 under *low* $C_{mt}$ and 0.083 under *medium* $C_{mt}$. Both values are significantly lower than the silent policy's mean of 0.160 and the Jennings' policy's mean of 0.161. Thus, we have been able to quantify the savings provided by STEAM over less selective policies within this example domain.

However, within a given domain configuration, STEAM must either always or never communicate, and this inflexibility leads to significant suboptimality across a wide range of domain configurations. On the other hand, Figure 6 also shows that there are domain configurations where STEAM is locally optimal. In this relatively small-scale experimental testbed, there is no need to incur STEAM's suboptimality, because the agents can compute the superior locally optimal policy in under 5 seconds. In larger-scale domains, on the other hand, the increased complexity of the locally optimal policies may render its execution infeasible. In such domains, STEAM's constant-time execution would potentially make it a preferable alternative. This analysis suggests a possible spectrum of algorithms that make different optimality-efficiency tradeoffs.

To understand the cause of STEAM's suboptimality, we can examine its performance more deeply in Figures 7 and 8, which plot the expected number of messages sent using STEAM (with both *low* and *medium* $C_{mt}$) vs. the locally optimal policy, at observability values of 0.3 and 0.7. STEAM's expected number of messages is either 0 or 1, so STEAM can make at most two (instantaneous) transitions between them: one threshold value each along the observability and communication cost dimensions.

From Figures 7 and 8, we see that the optimal policy can be more flexible than STEAM by specifying communication contingent on *Escort*'s beliefs beyond simply the achievement of $G_R$. For example, consider the messages sent under *low* $C_{mt}$ in Figure 7, where STEAM matches the locally optimal policy at the extremes of the communication cost dimension. Even if the communication cost is high, it is still worth sending message $\sigma_{G_R}$ in states where *Transport* is still very far from the destination. Thus, the surface for the optimal policy, makes a more gradual transition from always communicating to never communicating. We can thus view STEAM's surface as a crude approximation to the optimal surface, subject to STEAM's fewer degrees of freedom.

We can also use Figures 7 and 8 to identify the domain conditions under which joint intentions theory's prescription of attaining mutual belief is or is not optimal. In particular, for any domain where the observability is less than 1, the agents will not attain mutual belief without communication. In both Figures 7 and 8, there are *many* domain configurations where the locally optimal policy is expected to send fewer than 1 $\sigma_{G_R}$ message. Each of these configurations represents a domain where the locally optimal policy will not attain mutual belief in at least one case. Therefore, attaining mutual belief is suboptimal in those configurations!

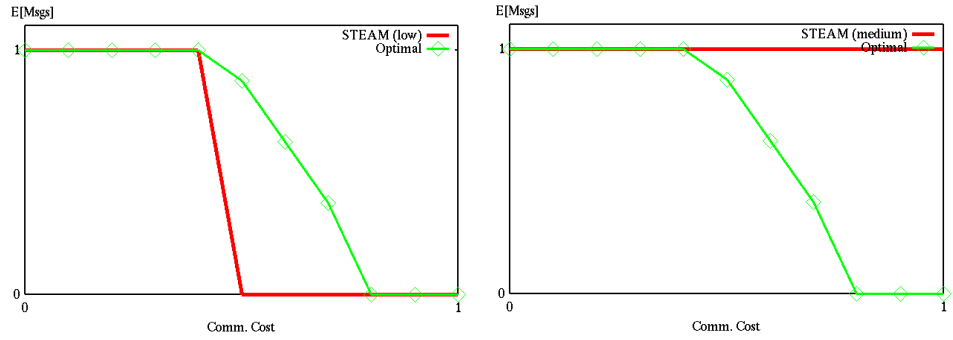

Figure 7: Expected number of messages sent by STEAM and locally optimal policies when the observability is 0.3.

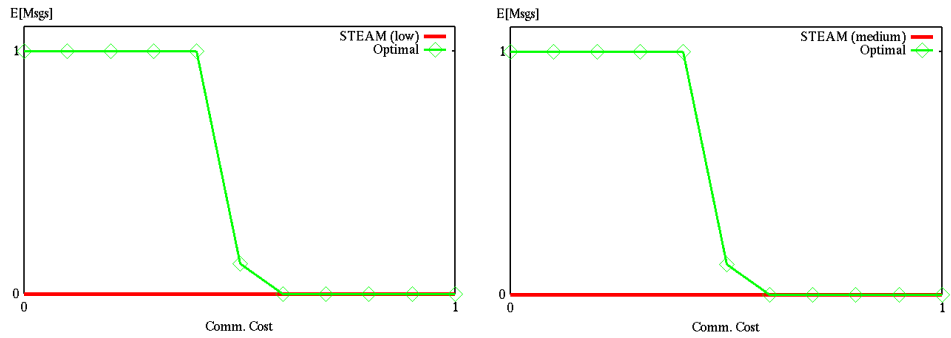

Figure 8: Expected number of messages sent by STEAM and locally optimal policies when the observability is 0.7. Under both settings, STEAM sends 0 messages.

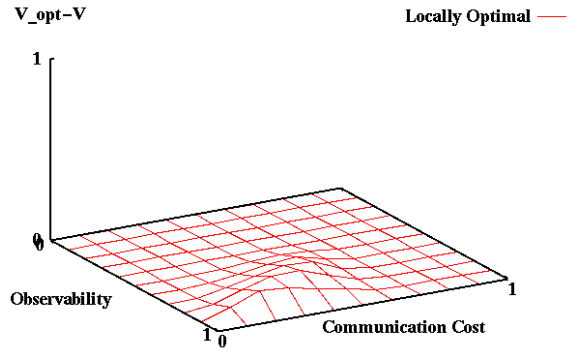

Figure 9: Suboptimality of locally optimal policy.

These experiments illustrate that STEAM, despite its decision-theoretic communication selectivity, may communicate suboptimally under a significant class of domain configurations. Previous work on STEAM-based, real-world, agent-team implementations informally noted suboptimality in an isolated configuration within a more realistic helicopter transport domain (Tambe, 1997). Unfortunately, this previous work treated that suboptimality (where the agents communicated more than necessary) as an isolated aberration, so there was no investigation of the degree of such suboptimality, nor of the conditions under which such suboptimality may occur in practice. We re-created these conditions within the experimental testbed of this section by using a *medium* $C_{mt}$. The resulting experiments (as shown in Figure 7) illustrated that the observed suboptimality was not an isolated phenomenon, but, in fact, that STEAM has a general propensity towards extraneous communication in situations involving low observability (i.e., low likelihood of mutual belief) and high communication costs. This result matches the situation where the "aberration" occurred in the more realistic domain.

The locally optimal policy is itself suboptimal with respect to the globally optimal policy, as we can see from Figure 9. Under domain configurations with high observability, the globally optimal policy has the escort wait an additional time step after destroying the radar and then communicate only if the transport continues flying nap-of-the-earth. The escort cannot directly observe which method of flight the transport has chosen, but it can measure the change in the transport's position (since it maintains a history of its past observations) and thus *infer* the method of flight with complete accuracy. In a sense, the escort following the globally optimal policy is performing *plan recognition* to analyze the transport's possible beliefs. It is particularly noteworthy that our domain specification does not explicitly encode this recognition capability. In fact, our algorithm for finding the globally optimal policy does not even make any of the assumptions made by our locally observable policy (i.e., single agent is deciding whether to communicate or not, regarding a single message, at a single point in time); rather, our general-purpose search algorithm traverses the policy space and "discovers" this possible means of inference on its own. We expect that such COM-MTDP analysis can provide an automatic method for discovering novel communication policies of this type in other domains, even those modeling real-world problems.

Indeed, by exploiting this discovery capability within our example domain, the globally optimal policy gains a slight advantage in expected utility over the locally optimal policy, with a mean difference of 0.011, standard deviation of 0.027, and maximum of 0.120. On the other hand, our domain-independent code never requires more than 5 seconds to compute the locally optimal policy in this testbed, while our domain-independent search algorithm always required more than 150 *minutes* to find the globally optimal policy. Thus, through Theorem 7, we have used the COM-MTDP model to construct a communication policy that, for this testbed domain, performs almost optimally and outperforms existing teamwork theories, with a substantial computational savings over finding the globally optimal policy. Although these results hold for an isolated communication decision, we expect the relative performance of the policies to stay the same even with multiple decisions, where the inflexibility of the suboptimal policies will only exacerbate their losses (i.e., the shapes of the graphs would stay roughly the same, but the suboptimality magnitudes would increase).

## 6. Summary

The COM-MTDP model is a novel framework that *complements* existing teamwork research by providing the previously lacking capability to analyze the optimality and complexity of team decisions. While grounded within economic team theory, the COM-MTDP's extensions to include communication and dynamism allow it to subsume many existing multiagent models. We were able to exploit the COM-MTDP's ability to represent broad classes of multiagent team domains to derive complexity results for optimal agent teamwork under arbitrary problem domains. We also used the model to identify domain properties that can simplify that complexity.

The COM-MTDP framework provides a general methodology for analysis across both general domain subclasses and specific domain instantiations. As demonstrated in Section 4, we can express important existing teamwork theories within a COM-MTDP framework and derive broadly applicable theoretical results about their optimality. Section 5 demonstrates our methodology for the analysis of a specific domain. By encoding a teamwork problem as a COM-MTDP, we can use the leverage of our general-purpose software tools (available in Online Appendix 1) to evaluate the optimality of teamwork based on potentially any other existing theory, as demonstrated in this paper using two leading instantiations of joint intentions theory. In combining both theory and practice, we can use the theoretical results derived using the COM-MTDP framework as the basis for new algorithms to extend our software tools, just as we did in translating Theorem 7 from Section 4 into an implemented algorithm for locally optimal communication in Section 5. We expect that the COM-MTDP framework, the theorems and complexity results, and the reusable software will form a basis for further analysis of teamwork, both by ourselves and others in the field.

## 7. Future Work for COM-MTDP Team Analysis

While our initial COM-MTDP results are promising, there remain at least three key areas where future progress in COM-MTDPs is critical. First, analysis using COM-MTDPs (such as the one presented in Section 5) requires knowledge of the rewards, transition probabilities, and observation probabilities, as well as of the competing policies governing agent behavior. It may not always be possible to have such a model of the domain and agents' policies readily available. Indeed, other proposed team-analysis techniques (Nair, Tambe, Marsella, & Raines, 2002b; Raines, Tambe, & Marsella, 2000), do not require a priori hand-coding of such models, but rather acquire them automatically through machine learning over large numbers of runs. Also, in the interests of combating computational complexity and improved understandability, some researchers emphasize the need for multiple models at multiple levels of abstraction, rather than focusing on a single model (Nair et al., 2002b). For instance, one level of the model may focus on the analysis of the individual agents' actions in support of a team, while another level may focus on interactions among subteams of a team. We can potentially extend the COM-MTDP model in both of these directions (i.e., machine learning of model parameters, and hierarchical representations of the team to provide multiple levels of abstraction).

Second, it is important to extend COM-MTDP analysis to other aspects of teamwork beyond communication. For instance, team formation (where agents may be assigned specific roles within the team) and reformation (where failure of individual agents leads to role reassignment within in the team) are key problems in teamwork that appear suitable for COM-MTDP analysis. Such analysis may require extensions to the COM-MTDP framework (e.g., explicit modeling of *roles*). Ongoing research (Nair, Tambe, & Marsella, 2002a) has begun investigating the impact of such extensions and their applications in domains such as RoboCup Rescue (Kitano, Tadokoro, Noda, Matsubara, Takahashi, Shinjoh, & Shimada, 1999). Analysis of more complex team behaviors may require further extensions to the COM-MTDP model to explicitly account for additional aspects of teamwork (e.g., notions of authority structure within teams).

Third, extending COM-MTDP analysis beyond teamwork to model other types of coordination may require relaxation of COM-MTDP's assumption of selfless agents receiving the same joint reward. More complex organizations may require modeling other non-joint rewards. Indeed, enriching the COM-MTDP model in this manner may enable analysis of some of the seminal work in multiagent coordination in the tradition of PGP and GPGP (Decker & Lesser, 1995; Durfee & Lesser, 1991). Such enriched models may first require new advances in the mathematical foundations of our COM-MTDP framework, and ultimately contribute towards the emerging sciences of agents and multiagent systems.

## Acknowledgments

This article is a significantly extended version of a paper, "Multiagent Teamwork: Analyzing the Optimality and Complexity of Key Theories and Models", by the same authors, in the *Proceedings of the International Joint Conference on Autonomous Agents and Multi-Agent Systems*, 2002. This article extends the initial content by providing proofs missing in the original paper, as well as new theoretical results, a detailed description of our experimental setup, new experimental results, and additional discussion and explanations of key points. This research was supported by DARPA award No. F30602-98-2-0108, under the Control of Agent Based Systems program, and managed by AFRL/Rome Research Site. We would like to thank Daniel Bernstein, Ashish Goel, Daniel Marcu, Stacy Marsella, Ranjit Nair, and Paul Rosenbloom for valuable discussion and feedback. We also thank the anonymous reviewers for their helpful comments and suggestions.